\pdfoutput=1

\documentclass[11pt]{article}

\usepackage[]{acl}

\usepackage{times}
\usepackage{latexsym}

\usepackage{adjustbox}
\usepackage{amsmath}
\usepackage{amssymb}
\usepackage{booktabs}
\usepackage{multirow}
\usepackage{todonotes}
\usepackage{enumitem}

\usepackage[T1]{fontenc}

\usepackage[utf8]{inputenc}

\usepackage{microtype}

\newcommand*{\courier}{\fontfamily{pcr}\selectfont}

\usepackage{mathtools}

\title{FAMuS: Frames Across Multiple Sources}

\newcommand{\ur}{\textsuperscript{\rm 1}}
\newcommand{\jhu}{\textsuperscript{\rm 2}}
\author{Siddharth Vashishtha\ur\,\,\,\,\, Alexander Martin\ur\,\,\,\,\, William Gantt\ur \\ \,\bf Benjamin Van Durme\jhu\,\,\,\,\,Aaron Steven White\ur\,\,\,\,\, 
\\
  \ur~University of Rochester~~\jhu~Johns Hopkins University~~\\
    \courier{\small\{svashis3@cs.|wgantt@cs.|amart50@u.|aaron.white@\}rochester.edu}
}

\begin{document}
\maketitle
\begin{abstract}
Understanding event descriptions is a central aspect of  language processing, but current approaches focus overwhelmingly on single sentences or documents. Aggregating information about an event \emph{across documents} can offer a much richer understanding. To this end, we present FAMuS, a new corpus of Wikipedia passages that \emph{report} on some event, paired with underlying, genre-diverse (non-Wikipedia) \emph{source} articles for the same event. Events and (cross-sentence) arguments in both report and source are annotated against FrameNet, providing broad coverage of different event types. We present results on two key event understanding tasks enabled by FAMuS: \emph{source validation}---determining whether a document is a valid source for a target report event---and \emph{cross-document argument extraction}---full-document argument extraction for a target event from both its report and the correct source article. We release both FAMuS and our models to support further research.
\end{abstract}

\section{Introduction}
\label{sec:introduction}
Recent years have witnessed a resurgence of interest in document-level event and argument extraction tasks, such as \emph{template filling} \citep{du-etal-2021-template, chen-etal-2023-iterative, gantt2022event}, \emph{role-filler entity extraction} \citep{du-etal-2021-grit, huang-etal-2021-document}, and \emph{event argument extraction} \citep{ebner-etal-2020-multi, li-etal-2021-document, tong-etal-2022-docee}. Indeed, the earliest goals of information extraction (IE), as advanced by the Message Understanding Conferences (MUCs), were to develop systems capable of extracting document-level event structures \citep{grishman-sundheim-1996-message, grishman_2019}. While the renewed interest in these goals represents clear progress beyond the longstanding and dominant focus on \emph{sentence-level} event extraction, recent work in this area suffers from two key shortcomings.

\begin{figure}
    \centering
    \includegraphics[width=\columnwidth]{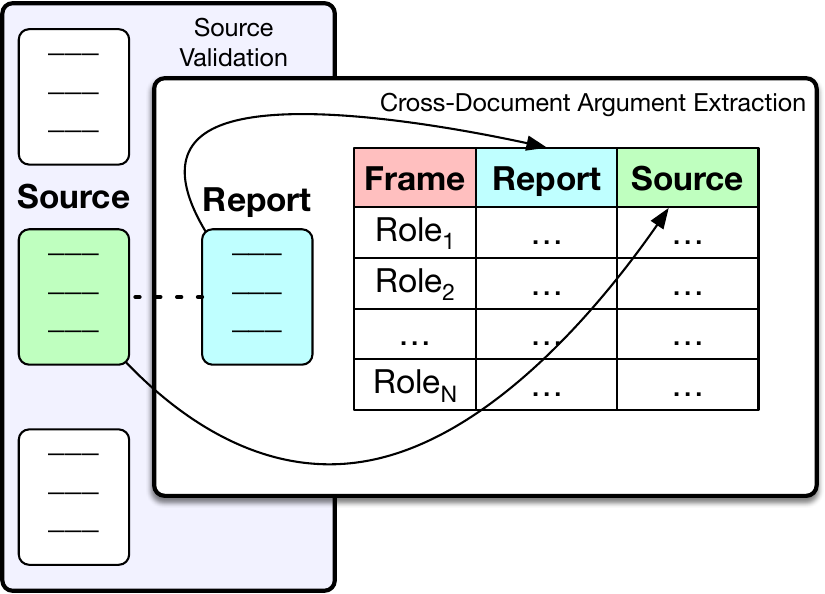}
    \caption{Schematic of the two FAMuS tasks: source validation and cross-document argument extraction.}
    \vspace{-1.5em}
    \label{fig:famus-schematic}
\end{figure}

For one, major benchmarks on these tasks, including MUC-4 \citep{muc-1992-message}, RAMS \citep{ebner-etal-2020-multi}, WikiEvents \citep{li-etal-2021-document}, and DocEE \citep{tong-etal-2022-docee} feature highly domain-specific event ontologies. Even when the absolute number of types is relatively large (e.g.\ the 139 event types covered by RAMS), they tend to be tightly clustered within a small handful of categories.

For another, although whole-document extraction enables a richer understanding of an event than its sentence-level analogue, it is still constrained by the input document's description of that event, which may lack key details. The task of \emph{event linking} partly remedies this by linking event mentions to a canonical entry in a knowledge base, but stops there, providing no actual extractions from those entries \citep{nothman-etal-2012-event, yu-etal-2023-event}.



\begin{figure*}[t]
    \centering
    \includegraphics[width=\textwidth]{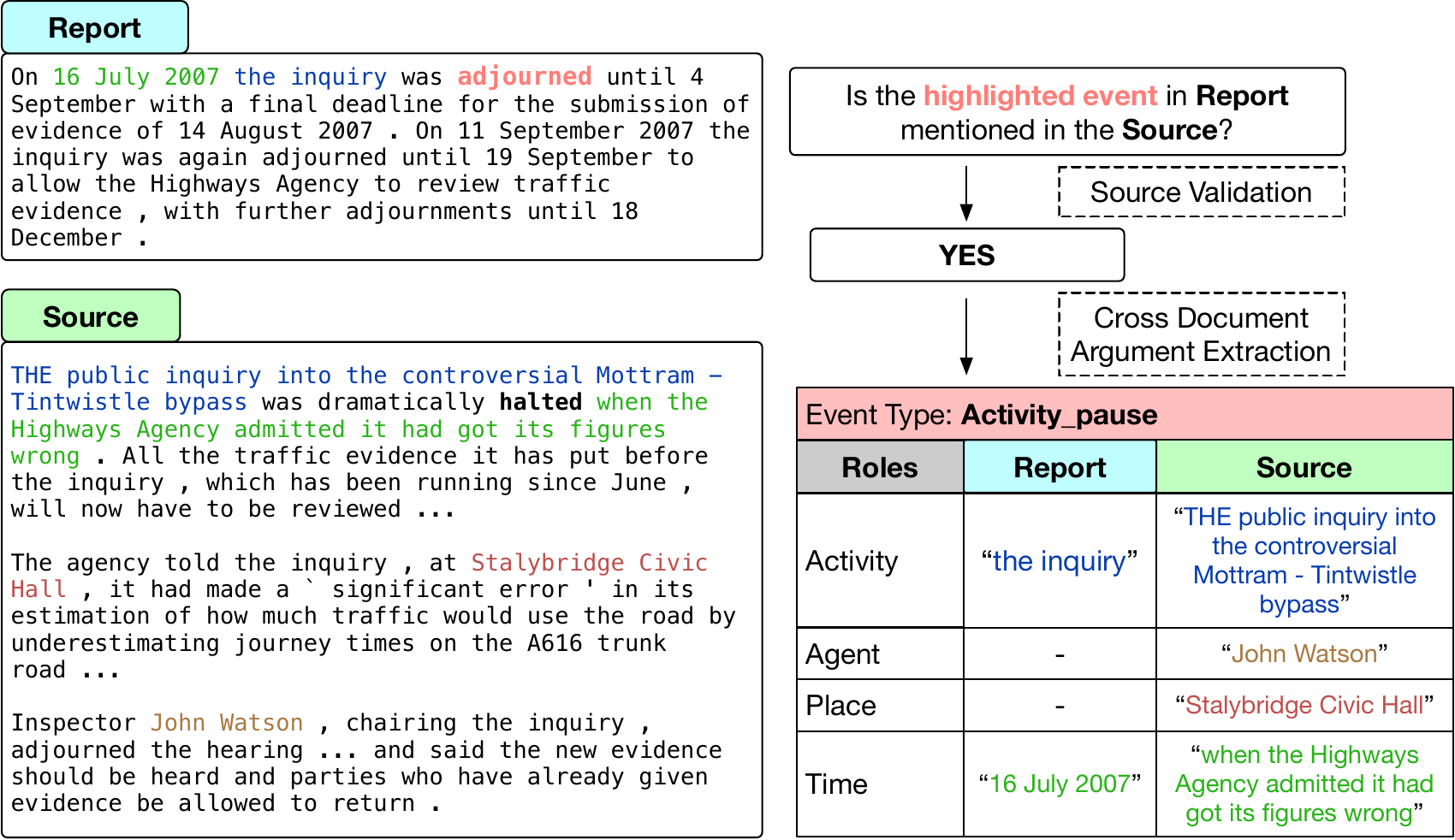}
    \caption{An example from FAMuS. The \textbf{Source Validation} task asks whether the event denoted by the trigger highlighted in the report text (\emph{adjourned}) is also described in the source text. If so, the system must then identify and extract all arguments of that event in \emph{both} the report and the source in the \textbf{Cross-Document Argument Extraction} task. FAMuS contains genre-diverse $(\text{report},\text{source})$ pairs selected from the MegaWika dataset \citep{barham2023megawika} and annotates a single target event trigger in the report, along with all arguments in both report and source, against FrameNet \citep{baker-etal-1998-berkeley}, enabling broad coverage of different event types.} 
    \vspace{-1em}
    \label{fig:famus-example}
\end{figure*}

This work introduces \textbf{FAMuS} (\textbf{F}rames \textbf{A}cross \textbf{Mu}ltiple \textbf{S}ources), a dataset and benchmark aimed at addressing both of these shortcomings. FAMuS provides event and cross-sentence argument annotations on over 1,255 Wikipedia passages (or \emph{reports}), each paired with cross-sentence argument annotations for the \emph{same} event as described in the document cited as the passage's \emph{source}. Events and arguments are annotated against FrameNet \citep{baker-etal-1998-berkeley}, providing genuinely broad coverage with 253 diverse event types and five supporting documents per type. Beyond the dataset itself, we make the following further contributions:
\begin{itemize}[itemsep=-0.25em]
    \item We introduce a novel cross-document objective (\autoref{fig:famus-schematic}), supported by FAMuS, comprising two challenging tasks: (i) \textbf{Source Validation}, which requires determining whether an input document is a valid \emph{source} for a tagged event in a given report; and (ii) \textbf{Cross-Document Argument Extraction}, which requires extracting arguments for a tagged report event from \emph{both} the report and its source.
    \item We present a results from a diverse suite of models on both tasks, including heuristic baselines, fine-tuned models using off-the-shelf encoders, and few-shot LLMs.
    \item We propose a new evaluation metric for argument extraction that computes an edit distance-based soft match between predicted and reference arguments to provide a richer picture of systems' argument extraction performance than traditional exact match.
\end{itemize}
The FAMuS dataset and baselines are available at \url{https://github.com/FACTSlab/FAMuS}.

\section{Task Definitions}
\label{sec:task_setup}
To situate FAMuS in the context of prior work, we first give a formal statement of the tasks it presents:
\begin{enumerate}[itemsep=-0.25em]
    \item \textbf{Source Validation (SV)}. Given a \emph{report text} $R$, a target event trigger (mention) $e$ occurring in $R$, and a candidate \emph{source text} $S$, determine whether $S$ contains a description of the same event as the one denoted by $e$.
    \item \textbf{Cross-Document Argument Extraction (CDAE)}. Given a report text $R$, a target event trigger $e$ in $R$, and a \emph{correct} source text $S$, extract all arguments of $e$ in both $R$ and $S$. We assume $e$ is assigned an event type from some underlying ontology of event types $E_1, \ldots E_N$, where each $E_i$ has roles $R^{(i)}_1, \ldots, R^{(i)}_{M_i}$, and where $e$'s arguments must each be assigned one of these roles.\footnote{Note that we do \emph{not} require $S$ to contain an explicit event trigger $e'$ coreferent with $e$. We require only that $S$ refers somehow to the event denoted by $e$, even if this reference is made more obliquely than with a single lexical item.}
\end{enumerate}
Both tasks are illustrated schematically in \autoref{fig:famus-schematic} and in greater detail in \autoref{fig:famus-example}. Taken together, these tasks are a more formal embodiment of \emph{informal} reading habits familiar to researchers and casual internet users alike: in the course of our reading, we learn of an event that intrigues us and then we seek out further information about it in other \emph{relevant} sources.

\section{Background}
\label{sec:background}
We are aware of no prior work that combines identification of a report event's source document (SV) with argument extraction from both the report and the source (CDAE). However, both closely relate to a number of established tasks in the literature, which we survey briefly below.

\paragraph{Event Linking (EL)} or \emph{event grounding} is the task of associating an event description (typically, a single mention) with a canonical entry for that event in some knowledge base. It resembles SV in attempting to ground a target event mention in a text to a more comprehensive description of the same event in a source text. But whereas SV takes a candidate source text as input (along with the report), EL aims to produce (a link to) one as output.

Introduced by \citet{nothman-etal-2012-event} as an event-centric analogue to the more popular \emph{entity linking} objective \citep{bunescu-pasca-2006-using, ji-grishman-2011-knowledge}, EL has received comparatively little attention. While \citeauthor{nothman-etal-2012-event} used Australian news articles for both report and source, more recent efforts have focused on Wikipedia and Wikidata. \citet{yu-etal-2023-event} use Wikipedia articles as source documents and present evaluations with both Wikipedia and New York Times report articles. \citet{ou2023hierarchical}, extending work by \citet{pratapa-etal-2022-multilingual}, propose an interesting hierarchical variant of the task, in which mentions must be linked to a \emph{set} of hierarchically related events in WikiData.

\paragraph{Cross-Document Event Coreference (CDEC)} involves identifying all coreferring event mentions across a collection of documents \citep{bagga-baldwin-1999-cross}. Various benchmarks exist for the task, including ECB+ \citep{cybulska-vossen-2014-using}, MEANTIME \citep{minard-etal-2016-meantime}, the Gun Violence Corpus \citep[GVC;][]{vossen-etal-2018-dont}, and WEC \citep{eirew-etal-2021-wec,eirew-etal-2022-cross}. From one angle, CDEC can be viewed as a kind of generalization of EL, insofar as the latter is concerned only with matching up \emph{pairs} of documents that describe the same event, and the former with matching up (potentially) multiple. However, CDEC usually expressly clusters event \emph{mentions}, whereas EL and SV often do not.

\paragraph{Claim Verification} SV is also structurally similar to  \emph{fact} or \emph{claim verification}, in which the goal is to determine whether some target statement (the \emph{claim}) is supported, unverified, or refuted by a source text.\footnote{In some cases, the relevant evidentiary sentences from the source must also be provided.} Notable benchmarks here include Emergent \citep{ferreira-vlachos-2016-emergent}, the Fake News Challenge \citep{pomerleau-rao-2017-fake}, LIAR \citep{wang-2017-liar}, and FEVER \citep{thorne-etal-2018-fever}. Although they are \emph{structurally} similar, the underlying relations governing each task (event coreference and evidentiary support) are clearly distinct.

\paragraph{Event Argument Extraction (EAE)} is a generalization of semantic role labeling \citep[SRL;][]{gildea-jurafsky-2002-automatic} that additionally assigns roles to a predicate's extra-sentential arguments.\footnote{EAE is synonymous with \emph{multi-sentence argument linking} and arguably also with \emph{implicit semantic role labeling}, though exact task definitions differ. See \citet{o2019bringing} and \citet{gantt2021argument} for surveys.} Our CDAE subtask is just EAE applied to both the report and the source texts. SemEval 2010 Task 10 \citep{ruppenhofer-etal-2010-semeval} and Beyond NomBank \citep{gerber-chai-2010-beyond} represent the first true benchmarks for EAE, with the former consisting of a set of Sherlock Holmes stories annotated against FrameNet, and the latter annotating the arguments of a set of 10 nominal predicates from NomBank \citep{meyers-etal-2004-nombank} on the Penn Tree Bank corpus \citep{marcus1993building}. Other resources include ONV5 \citep{moor2013predicate} and MS-AMR \citep{ogorman-etal-2018-amr}. Unfortunately, these datasets are all quite small: the largest, MS-AMR, still contains only about 2,400 implicit arguments. EAE has lately seen renewed interest due mainly to the much larger RAMS \citep{ebner-etal-2020-multi} and WikiEvents \citep{li-etal-2021-document} benchmarks (20-30k arguments each). The more recent DocEE \citep{tong-etal-2022-docee} benchmark is an order of magnitude larger still (180k arguments). One disadvantage of these three datasets relative to their predecessors, however, is their use of domain-specific ontologies. FAMuS aims to address both of the above issues by providing a relatively \emph{large} dataset annotated against a \emph{broad-coverage} ontology.

\paragraph{Predicate-Argument Alignment} Related to CDAE (and CDEC), some prior work has studied cross-document alignment of predicate-argument structures. \citet{roth-frank-2012-aligning}, for instance, annotate gold predicate alignments in 70 pairs of topically related documents from GigaPairs \citep{roth-frank-2012-aligning-predicate} and introduce a graph-based clustering model for the task. \citet{wolfe-etal-2013-parma} present PARMA, a feature-rich, regularized logistic regression model for the same task that makes independent alignment decisions for each predicate and argument. \citet{wolfe-etal-2015-predicate} then extend this work with a model that relaxes the independence assumption and \emph{jointly} models predicates and arguments. While CDAE demands neither identification of a predicate in the source document nor an explicit argument-to-argument alignment, it is similar to this work in identifying aligned \emph{sets} of arguments of the same event across documents.





\section{Data Collection}
\label{sec:data_collection}
The FAMuS documents represent a subset of English documents from MegaWika \citep{barham2023megawika}, a dataset comprising millions of (report, source) pairs across 50 languages. Below, we discuss data collection for our SV and CDAE tasks.

\begin{table}
    \centering
    \adjustbox{max width=\columnwidth}{
    \begin{tabular}{l|cc}
    \toprule
         & Train & Dev \\
    \midrule
        Event Types & 253 & 253 \\
        Role Types (R) & 712 & 580 \\
        Role Types (S) & 749 & 643 \\
        SV Examples ($+$) & 759 & 253 \\
        SV Examples ($-$) & 759 & 253 \\
        Avg. Tokens (R) & 59 & 60 \\
        Avg. Tokens (S) & 1,084 & 1,511 \\
        Avg.\ Filled Roles (R) & 2.97 & 3.45 \\
        Avg.\ Filled Roles (S) & 3.45 & 3.89 \\
        Avg.\ Args (R) & 3.07 & 3.55 \\
        Avg.\ Args (S) & 3.70 & 4.28 \\
    \bottomrule
    \end{tabular}
    }
    \caption{Summary statistics for the FAMuS train and dev splits (test deliberately omitted). ``(R)'' and ``(S)'' denote \emph{report} and \emph{source}, respectively. Note that CDAE examples (not shown) are the same as ``SV Examples ($+$),'' as these consist of the same documents (see \S\ref{sec:data_collection}).}
    \vspace{-1.2em}
    \label{tab:summary-stats}
\end{table}

\subsection{Source Validation}\label{subsec:source-validation}
\paragraph{Overview} Since source texts may be arbitrary web pages, it is essential to verify their quality \emph{as sources} for the report text. To this end, \citet{barham2023megawika} devise their own source validation task for (report, source) pairs in MegaWika, in which annotators on Amazon Mechanical Turk are presented with a highlighted event trigger in the report text and are asked two questions:
\begin{enumerate}[itemsep=-0.25em]
    \item What is the most likely FrameNet frame for the highlighted text in the report?\footnote{Annotators are shown the top five candidate frames from a FrameNet parser along with a ``none'' option.}
    \item {Does the source describe the same event as is denoted by the highlighted report trigger?}
\end{enumerate}
Each pair is evaluated by three annotators. The authors observe low inter-annotator agreement (IAA) on (2), obtaining a Krippendorff's $\alpha$ of 0.41 \citep{krippendorff2018content}. Furthermore, taking the majority response across annotators on (2) reveals that only about 48\% of source documents actually describe the same event as the associated report.

In this work, we refine \citeauthor{barham2023megawika}'s methodology for our own SV annotation. To ensure high quality annotations, we restrict \emph{positive} SV examples to the set of (report, source) pairs where (i) a majority (2/3) agrees on the report event's correct frame, and (ii) \emph{either} all three annotators \emph{unanimously} agree that the source is a valid one for the report event \emph{or} 2/3 agree and an expert (one of the authors) agrees with the majority.\footnote{A subset of the authors inspected all 2/3 majority cases.} We then identify negative examples (i.e.\ reports with \emph{incorrect} sources) using a combination of manual and automatic methods. We detail the methodology for both positive and negative example curation below.

\paragraph{Positive Examples}
A key consideration in selecting positive examples is to ensure \emph{broad coverage} of events---both in terms of the number of different event \emph{types} and in terms of the number of examples for each type. Under a limited annotation budget, these goals trade off with each other, where more examples per type means reduced resources for annotation of additional types. Our data collection procedure targets a reasonable compromise between these aims. At a high level, we rely on the FrameNet inheritance hierarchy to identify a subset of 328 frames that denote a \emph{situation}---i.e.\ an \textsc{event}, \textsc{state}, or \textsc{process} in FrameNet.\footnote{Details on the frame selection process are in \autoref{app:frame_selection}.} We then iterate \citeauthor{barham2023megawika}'s annotation protocol until we obtain at least \emph{five} (report, source) pairs per frame that satisfy our two criteria---(i) and (ii) above---for positive examples, for at least 75\% (250) of the 328 situation-denoting frames.

To select (report, source) pairs for annotation, we rely on the oversampling technique from \citet{barham2023megawika}, which leverages the LOME FrameNet parser of \citet{xia-etal-2021-lome} and a simple Longformer-based model for the SV task (see \S\ref{sec:experiments}). Broadly, for each frame, we want to estimate how many \emph{total} examples we need to annotate in order obtain five \emph{positive} ones. This count ($X$) is modeled as a negative binomial random variable $X \sim \text{NB}(r,p)$ with $r = 5$ denoting the desired number of positive examples and $p$ denoting the probability of an example being positive. Given our two criteria for positive examples, $p$ can be expressed as $p = P_i \cdot v$, where $P_i$ is the precision of the parser on frames of type $f_i$ and where $v$ is the test set accuracy of our SV model.\footnote{$P_i$ and $v$ correspond to our models' ability to correctly answer questions (1) and (2) above, respectively.} For frames for which the FrameNet test set has poor support ($<10$ examples), we use the \emph{average} precision across all frames, $P_\text{avg}$, in lieu of $P_i$, for a more robust estimate. Thus, the \emph{expected} number of examples needed to obtain five positive ones, $D_i = \mathbb{E}[X]$, is:



\vspace{-1em}
\begin{align}
D_i = 
\begin{cases}
  \lceil\frac{5}{P_i*v}\rceil, & \text{if }  count(f_i) \geq 10 \\
  \lceil\frac{5}{ P_{avg}*v}\rceil, & \text{otherwise}
\end{cases}
\end{align}
While annotating $D_i$ examples for frame $f_i$ will yield five positive examples \emph{in expectation} for $f_i$, multiple rounds of annotation are needed to actually obtain five positive examples for all frames. In total, we conducted seven rounds. In each round, for each frame $f_i$, we use stratified sampling to ensure diversity among the $D_i$ (report, source) pairs selected for annotation. We first identify a candidate set $c_i$ of 250 pairs from MegaWika for which the FrameNet parser has identified at least one instance of frame $f_i$ in the report.\footnote{If fewer than 250 pairs are available for $f_i$, we include all $N_i < 250$ pairs.} We then perform $k$-means clustering on all pairs, clustering on the SpanBERT \citep{joshi-etal-2020-spanbert} \texttt{CLS} token embedding of the first five sentences of the source text for each pair, fixing $k = D_i$. We then sample one pair from each cluster, aiming to select pairs for which the report trigger's lemma differs from those in all other pairs chosen for $f_i$.

\begin{figure}[t]
    \centering
    \small
    \includegraphics[width=\columnwidth]{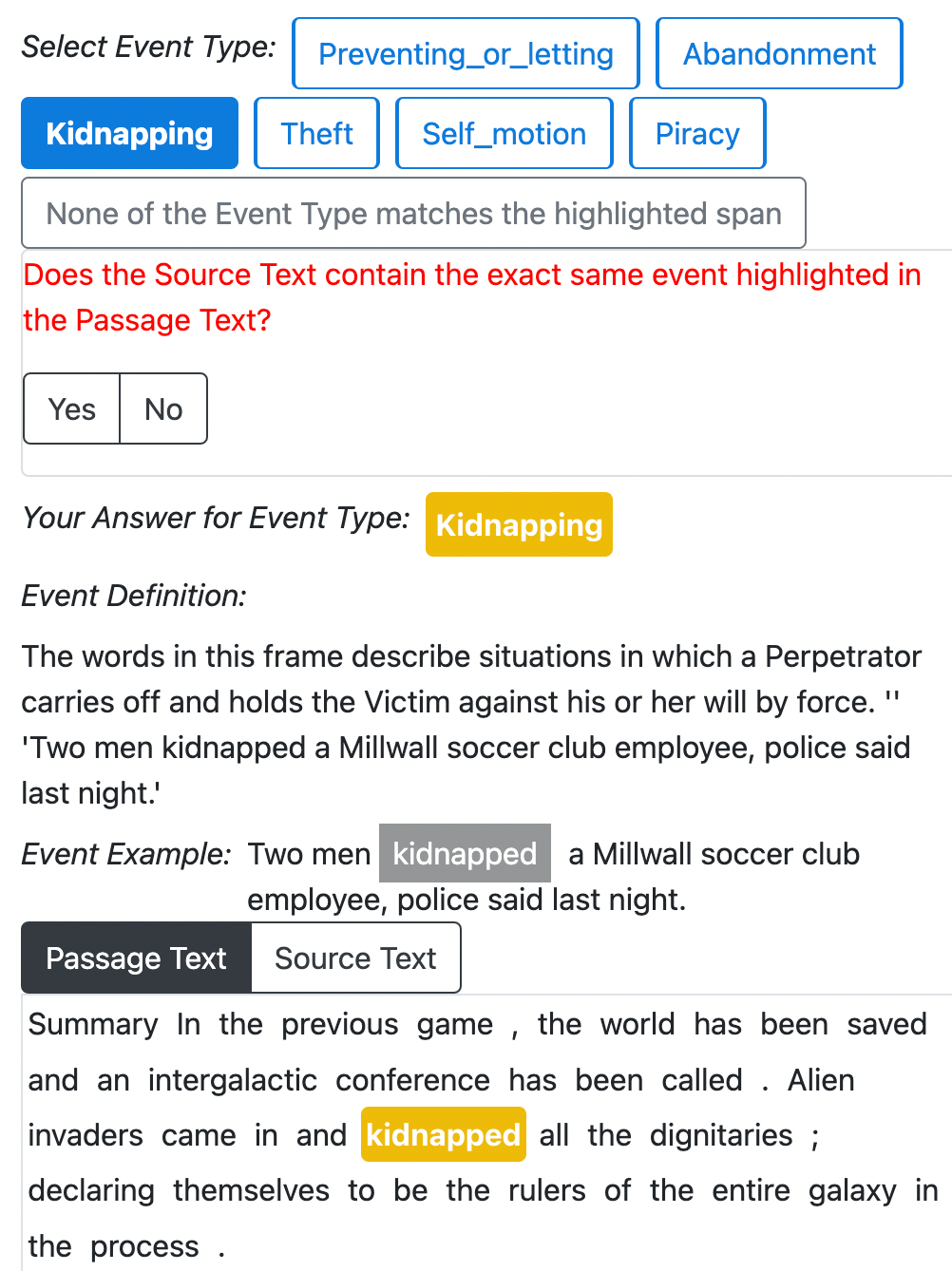}
    \caption{The source validation annotation interface, with the report (``passage'') text displayed. Annotators are shown the report with a highlighted event trigger and are asked to select the correct frame for the trigger from among the top five predictions of a FrameNet parser (or none, if all candidates are wrong). When a candidate frame is selected, its definition and an example from FrameNet are displayed.}
    \vspace{-1em}
    \label{fig:report-frame-val}
\end{figure}

\begin{figure}[t]
    \centering
    \small
    \includegraphics[width=\columnwidth]{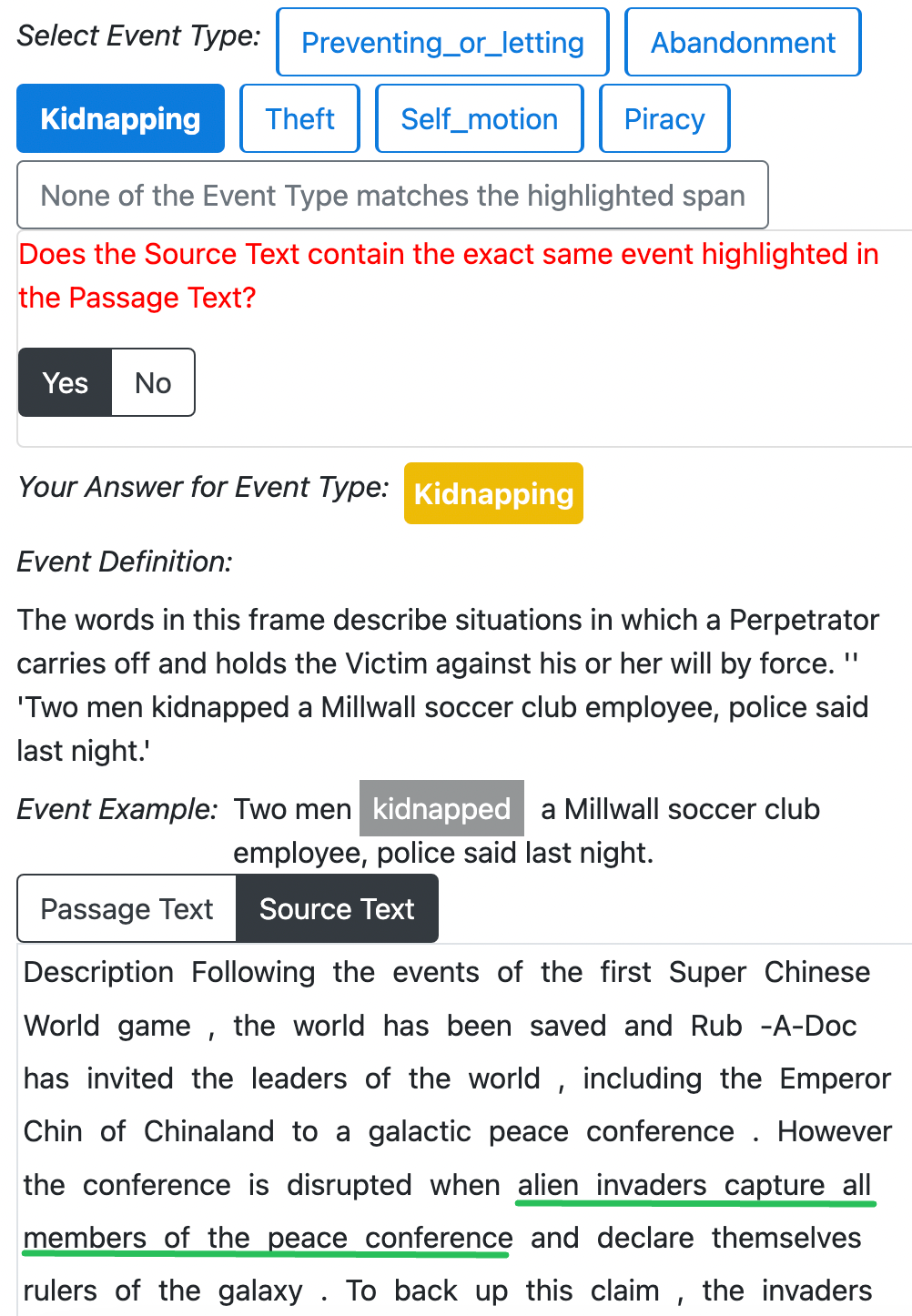}
    \caption{The same example as in \autoref{fig:report-frame-val}, but with (a portion of) the source text displayed. Here, the source document \emph{does} describe the same report event (relevant text underlined in green) as shown in \autoref{fig:report-frame-val}, and so is a valid source. Role annotation (\S\ref{subsec:role-annotation}) is done only on examples with valid source texts.}
    \vspace{-1em}
    \label{fig:source-frame-val}
\end{figure}

\autoref{fig:report-frame-val} shows an example of the source validation annotation interface with the report text displayed. \autoref{fig:source-frame-val} shows the same example, but with the \emph{source} text displayed, highlighting that the document \emph{is} a valid source for the report event.

\begin{figure}[t]
    \centering
    \small
    \includegraphics[width=\columnwidth]{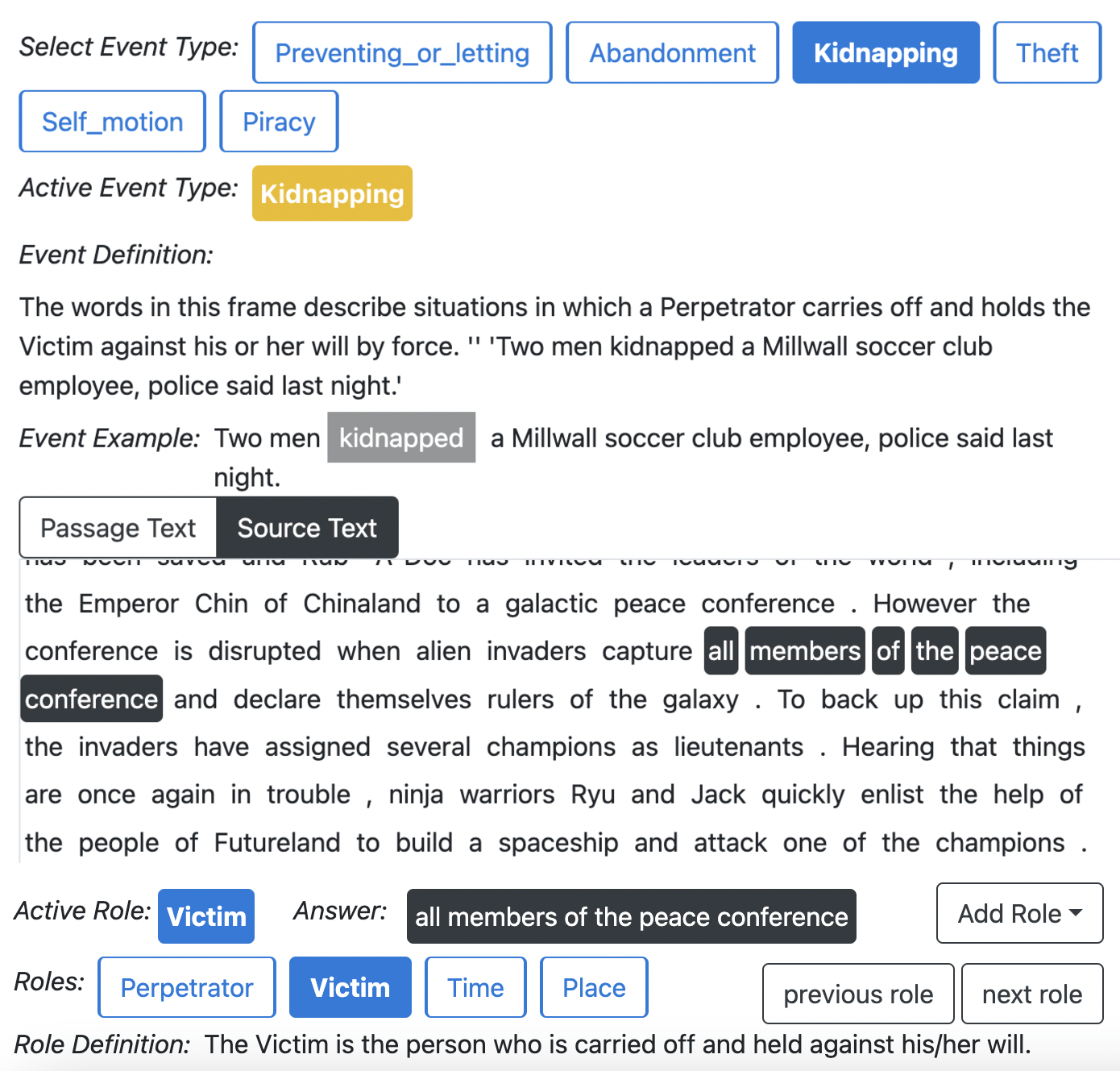}
    \caption{The role annotation interface for the same example as in \autoref{fig:report-frame-val}. Here, annotators identify arguments of the highlighted report (``passage'') event in the full texts of \emph{both} the report and source.\vspace{-5mm}}
    \label{fig:report-role-ann}
\end{figure}

\paragraph{Negative Examples}
To create a balanced dataset for the SV task, we also collect five negative examples for each frame. We take as many of these examples as possible from the annotated documents described above, subject to the criterion that all annotators unanimously agree that the source text \emph{does not} describe the same event as the report.

However, as the annotation protocol is designed mainly to identify positive examples, some frames still lacked five negative examples satisfying the above criterion at the end of the seven rounds. To make up the difference, we generate additional \textit{silver} negative examples automatically, as needed. We ensure that every example in the test set is either a gold example or manually annotated by one of the authors (platinum annotation) of this paper if it is a generated one.\footnote{11 examples in the test set were platinum annotated.}


Broadly, our approach entails selecting unannotated reports and pairing them with a \emph{new} source text that \emph{does not} describe the same event, but that is still semantically similar to the report, so as to ensure the task remains challenging. For each frame $f_i$, we start with the same candidate example set $c_i$ as described above, excluding all annotated examples to yield a new set $c'_i \subset c_i$. We then randomly select a pair $(r_i^{(j)}, s_i^{(j)})$ from $c'_i$. Next, we select the pair $(r_i^{(k)}, s_i^{(k)})$ from $c_i' - \{(r_i^{(j)}, s_i^{(j)})\}$ whose source ($s_i^{(k)}$) has highest similarity with $r_i^{(j)}$, forming a new negative example $(r_i^{(j)}, s_i^{(k)})$, where $j \neq k$ by construction. The similarity scores between a report and a source are obtained using SimLM, a state-of-the-art passage retrieval model \citep{wang-etal-2023-simlm}. While it is possible in principle that $s_i^{(k)}$ describes the same event as $r_j^{(k)}$, this is very unlikely in practice, given the massive pool of source documents we are drawing from. Additionally, distributions of report-source similarity scores for the positive examples and for these ``silver'' negative examples shown in \autoref{app:dataset_statistics} are quite divergent, underscoring this point. 


\paragraph{Annotation Quality} 
We implemented a two-stage qualification phase for Amazon Mechanical Turk workers involved in the CDAE task, as detailed in Section \ref{subsec:role-annotation}. All workers who qualified for the second phase were subsequently incorporated into the source validation task. In later iterations, an additional 11 workers, who achieved perfect agreement with the majority response of ten previously annotated gold-standard source validation examples, were included. This procedure resulted in a total of 26 workers dedicated exclusively to the source validation task, with each instance being annotated by three distinct workers. Each Human Intelligence Task (HIT) featured a single report-source pair and carried a reward of $0.20$ per HIT.

For frame identification (question (1)), we obtain Krippendorff's $\alpha$ of 0.62 across all annotated examples, indicating agreement consistent with---and in some cases, substantially better than---similar crowd-sourced frame identification tasks \citep[][\emph{i.a.}, see also \citealt{trott-etal-2020-construing}]{hong-baker-2011-good, fossati-etal-2013-outsourcing, vossen2018towards, vossen-etal-2020-large, dumitrache2018capturing}. For validation (question (2)), \emph{by construction}, all examples are either ones on which there was unanimous agreement, or else (for positive examples) had majority agreement, plus approval from one of the authors.

\subsection{Cross-Document Argument Extraction}\label{subsec:role-annotation}
\paragraph{Overview} In each round, after SV annotation, we collect \emph{full-document}\footnote{In contrast to much prior work on EAE, we do not impose fixed-size context windows during annotation, allowing arguments to be annotated \emph{anywhere} in the document.} role annotations on both the report and the (valid) source for the annotated report event. We annotate only the core roles of each frame, plus \textsc{Time} and \textsc{Place}. An example of the annotation interface is shown in \autoref{fig:report-role-ann}. Here, annotators select roles from the role set for the report trigger's frame and then select a contiguous span from the report or source text as an argument for that role. The interface also supports annotating multiple arguments for the same role. When a role is selected, its FrameNet definition and an example are displayed. Annotators are strongly encouraged to annotate based on the highlighted frame (chosen during SV annotation) but are permitted to change the frame in the rare case they deem it incorrect. Of the 1,255 CDAE examples annotated, only 4.6\% actually had their frame types changed---a testament to the high quality of the SV frame annotations.

While we do \emph{not} annotate for coreference, we do provide model-predicted (\emph{silver}) coreference clusters for all annotated arguments, which are used in one evaluation setting (see \S\ref{sec:experiments}). We use F-COREF \citep{otmazgin-etal-2022-f} as the coreference model.

\paragraph{Annotation Quality} Two qualifying phases were launched on Amazon Mechanical Turk (AMT) to select workers for CDAE annotation. In the first phase, annotators had to provide CDAE annotations for an example with a short ($\sim 250$-token) source document. In the second phase, annotators had to do the same, but for an example with a longer ($\sim 4$k-token) source document. Two authors of this paper performed the annotations for both phases and qualified annotators based on manual inspection of their work. Only workers who qualified in the first phase were allowed to participate in the second phase. Following qualification in the second phase, there were 15 total annotators selected for the main CDAE annotation. Annotators were paid \$1 for each task, and could receive a bonus of up to \$4 for especially high-quality annotations.

To ensure annotation quality remained high throughout data collection, we relied on a combination of automatic checks and manual correction. After the first round of annotation, two of the authors manually verified and corrected all annotations. The uncorrected data was then evaluated against the corrected data using the agreement metric described in \autoref{app:agreement}. Following the initial iteration, we observed a robust mean agreement for the report annotations, evidenced by a 0.94 F1 mean score. Simultaneously, the source annotations exhibited a mean 0.92 F1 score. Remarkably, the majority of these annotations achieved a perfect agreement score when compared with the corrected data.
Since manually correcting \emph{all} annotations ourselves was infeasible, we employed a hybrid (automatic/manual) verification method for rounds after the first. For each round, we computed agreement between ChatGPT-predicted annotations and the (uncorrected) Turker annotations, then manually corrected only examples that fell in the bottom quartile of agreement scores.\footnote{Details in \autoref{app:agreement}.} We also discarded a small number of examples due to poor document quality (e.g.\ large amounts of non-English text).

\section{Experiments}
\label{sec:experiments}
We now describe the models and setup for experiments on SV and CDAE. Model hyperparameters and prompts can be found in \autoref{app:model_details}.

\subsection{Source Validation}
Per \S\ref{sec:task_setup}, SV is a binary classification task that takes as input a report $R$, a (typed) event trigger $e \in R$, and a candidate source text $S$, and outputs a binary judgment indicating whether $S$ contains a description of the same event as is denoted by $e$. We consider three models for this task, in addition to a majority-class baseline.

\paragraph{Lemma Baseline} This model simply predicts \texttt{YES} if the lemma of the report's event trigger exists in the (lemmatized) source, and \texttt{NO} otherwise. We use NLTK's \texttt{WordNetLemmatizer} to obtain lemmas \citep{bird2009natural}.

\paragraph{Longformer} We use Longformer \citep{beltagy2020longformer} with a classification head and fine-tune it on FAMuS. The input sequence to the Longformer model is a \texttt{</s>}-delimited concatenation of the report and source text, with the report event's trigger marked by \texttt{<event>} tags.\footnote{The SV model we use for oversampling (see \S\ref{sec:data_collection}) is the same, except that we fine-tune it on \citeauthor{barham2023megawika}'s SV data.}

\paragraph{Few-Shot LLMs} Inspired by the successes of recent large language models (LLMs) on many IE tasks \citep{wei2023zero}, we also evaluate ChatGPT (\texttt{gpt-3.5-turbo-0301}) and Llama 2 \citep[\texttt{llama-2-13b};][]{touvron2023llama} on FAMuS in the few-shot setting. The prompt (which is the same for both models) describes the task and includes two positive and two negative examples handwritten by one of the authors. We set model temperature to 0 to ensure consistent generations.

\begin{figure*}[t]
    \centering
    \tiny
    \includegraphics[width=\textwidth, scale=1]{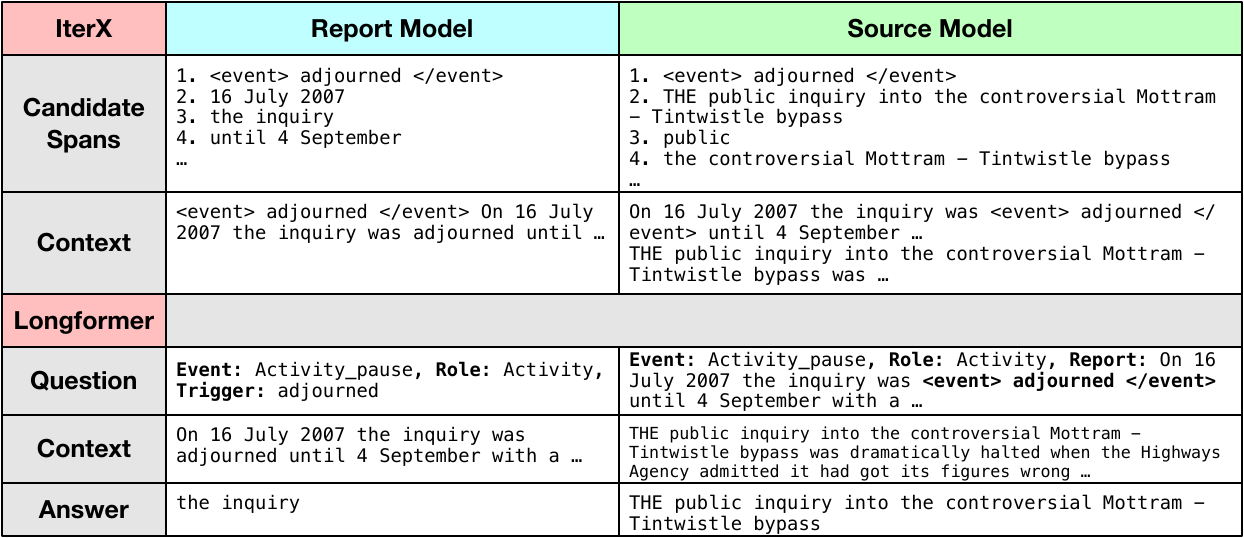}
    \caption{\textbf{Top}: IterX inputs for the example in \autoref{fig:famus-example}, including the set of candidate spans (first row) and the document text (second row) for the report (left) and source (right) models. \textbf{Bottom}: Longformer QA report (left) and source (right) model inputs for the \texttt{Activity} role for the example in \autoref{fig:famus-example}. Note that the question for the source model has the report text prepended, with the event trigger highlighted. This is done to condition extraction specifically on that trigger. See \S\ref{sec:experiments} for details.}
    \vspace{-3em}
    \label{fig:model_inputs}
\end{figure*}

\subsection{Cross Document Argument Extraction}
In CDAE, the input is a valid (report, source) pair, along with a (typed) event trigger in the report. The output is a set of arguments for the trigger, extracted from both report and source. Below, we present results on three CDAE models, training and evaluating each separately on report and source.

\paragraph{IterX} \citep{chen-etal-2023-iterative} is a recent template filling model that has achieved state-of-the-art performance on the MUC-4 \citep{sundheim-1992-overview, muc-1992-message} and SciREX \citep{jain-etal-2020-scirex} benchmarks. IterX treats template prediction as autoregressive span assignment, assigning each of a set of candidate input spans to a role (possibly null) in the current template, updating candidate embeddings based on those assignments, predicting another template, and so on, until all candidates are assigned the null role. Although IterX is designed to extract multiple templates per document, CDAE requires just one template each for the source and the report, containing all arguments of the input event. As such, we constrain IterX to decode a single template per document (report or source).


IterX requires the set of candidate spans to be given in the input. Depending on the setting (see below), these spans are drawn from different subsets of the following three sources: (i) gold spans obtained from CDAE annotation; (ii) argument spans extracted by the LOME FrameNet parser; (iii) entity spans identified by Stanza's NER module \citep{qi-etal-2020-stanza}. We consider three settings for training and evaluation: \textbf{gold spans}, where we train and test on (i); \textbf{predicted spans}, where we train on (i)-(iii) and test on (ii) and (iii); and \textbf{gold and predicted spans}, where train and test on (i)-(iii). Following standard practice in IE, these different settings are intended to show the relative value of access to gold spans during both training and inference.

IterX also takes as input an embedding for the type of template(s) to be generated. However, in contrast to templates in other datasets (e.g.\ MUC-4), FAMuS templates (i.e.\ frames) have lexical triggers, given as part of the CDAE input. Accordingly, we condition template prediction on both frame type \emph{and} trigger. We do this in two ways.

First, we always include the trigger (enclosed in \texttt{<event>} tags) among the input spans, but always assign it the special null role ($\epsilon$). Since IterX jointly encodes all input spans with a Transformer encoder \citep{vaswani2017attention}, the self-attention mechanism allows the role assignment decision for each span to be conditioned on the trigger.

Second, we modify the document text to highlight the trigger. For the report models, we simply prepend the \texttt{<event>}-enclosed trigger span to the report text. For the source models, we prepend the full report text (with \texttt{<event>}-enclosed trigger) to the source text. Example inputs are shown in \autoref{fig:model_inputs}. Following \citeauthor{chen-etal-2023-iterative}, we use IterX with a T5-large encoder \citep{raffel2020exploring}.



\paragraph{Longformer QA}
Our second model recasts CDAE as extractive question answering (QA) in the style of SQuAD 2.0 \citep{rajpurkar-etal-2018-know}, following much recent work in IE that takes a QA-based approach \citep[\emph{i.a.}]{du-cardie-2020-event, liu-etal-2020-event, holzenberger2022asking}.

We map each possible role of each report trigger's FrameNet frame, together with that role's gold argument(s), to a single QA pair. Separate QA datasets are created for the source and report annotations. For the report dataset, the context passage for each QA pair is the report text, the ``question''\footnote{The ``questions'' in QA-recasted IE datasets are often not syntactically interrogative \citep{du-cardie-2020-event}; we follow this looser notion of a question here.} is the concatenated names of the event and role, and the answer is the gold report argument(s) for that event and role. For the source dataset, the context passage is the source text; the question is the same as in the report model, but with the full report text (with marked event trigger) concatenated at the end; and the answer is again the gold source argument(s) for the given event and role. Examples for both QA datasets are shown in \autoref{fig:model_inputs}.

\paragraph{Few-Shot LLMs}
As with SV, we present few-shot evaluations on CDAE using Llama and ChatGPT in the few-shot setting.\footnote{We use \texttt{llama-2-13b-chat} (not \texttt{llama-2-13b}, as in SV). The ChatGPT version is unchanged.} The prompt (again, the same for both models) describes the task and includes two examples from the FAMuS training split, each consisting of a document (report or source) and its CDAE annotations.

\paragraph{Report Baseline} Finally, for the source document only, we report the score one would obtain by simply predicting the set of gold arguments from the \emph{report} document. Because this baseline uses (gold) arguments derived exclusively from the report, it isn't directly comparable with other results presented. Rather, it is mainly intended to give a quantitative sense of how much information about a situation---over and above the information in a report---sources tend to provide. Insofar as sources provide substantial additional information, we expect the results of this baseline to be poor.

We additionally report versions of all of the above models \emph{ensembled} with the report baseline ($+$rb) in Tables \ref{tab:cross_doc_arg_results} and \ref{tab:cross_doc_arg_with_silver_coref_results}: for any role $r$ in the target frame for which a model predicts no arguments, the $+$rb variant augments that model's predictions with all report arguments for role $r$, leaving predictions for all roles $r' \neq r$ unchanged.

\paragraph{Evaluation} For CDAE, we report the CEAF-RME metric of \citet{chen-etal-2023-iterative}. This metric  generalizes standard argument P/R/$\text{F}_1$ to allow for evaluation of models, like ours, that predict individual argument \emph{mentions} against references that have full entity coreference annotations.\footnote{The metric treats predicted mentions as singleton entities and allows multiple predicted singletons to be aligned to a given reference entity. We refer the reader to \citet{chen2023unified} and \citet{chen-etal-2023-iterative} for further details.} We report two versions of this metric. The first is \citeauthor{chen-etal-2023-iterative}'s $\text{CEAF-RME}_{\phi_3}$, which awards full credit to any predicted argument mention $p_r$ that \emph{exactly} matches some mention $g_r$ in a reference argument entity $C_{g_r}$ for the same role $r$.

The second version aims to accommodate the potentially significant variability in argument span boundaries,\footnote{For instance, annotators often mark NPs differently---e.g.\ some tend to include quantificational determiners, like \emph{every} and \emph{some}, or relative clauses while others do not.} employing a softer notion of span similarity than exact match, based on the normalized edit distance ($\hat{A}$) between $p_r$ and $g_r$:

{\scriptsize
\begin{align}\label{eq:a_hat}
\hat{A}(p_r, g_r) = 1 - \frac{\text{E}(p_r, g_r)}{(\text{S} - 1)\min(L_{p_r}, L_{g_r}) + \max(L_{p_r}, L_{g_r})}
\end{align}
}

\noindent where $E$ is Levenshtein distance with a substitution cost ($S$) of 2, and where $L_{p_r}$ and $L_{g_r}$ denote the number of tokens in $p_r$ and $g_r$. Instead of requiring exact match between $p_r$ and some mention in $g_r \in C_{g_r}$, we simply take the maximum $\hat{A}$ value between $p_r$ and any $g_r \in C_{g_r}$:
\vspace{-0.3em}
\begin{align}\label{eq:a}
a = \mathop{\mathrm{max}}\limits_{g_r \in C_{g_r}} \hat{A} (g_r, p_r)
\end{align}

\noindent We denote this second version as $\text{CEAF-RME}_a$. We report each version of the metric in two settings: one where we use only the (single) gold-annotated mention for each reference argument (\autoref{tab:cross_doc_arg_results}) and one where we use its full coreference cluster as predicted by F-COREF (\autoref{tab:cross_doc_arg_with_silver_coref_results}).

\section{Results}
\label{sec:results}
\begin{table}
    \centering
    \adjustbox{max width=\linewidth}{
    \begin{tabular}{lcccc}
    \toprule
    \textbf{Model} &  \textbf{Accuracy} & \textbf{P} & \textbf{R} & \textbf{F1}\\
    \midrule 
    Majority & 50.00 & 100.00 & 50.00 & 66.66 \\
    Lemma & \textbf{75.89} & 89.70 & 58.50 & 70.81 \\
    Longformer & 71.94 & 66.67 & \textbf{87.75} & \textbf{75.77}\\
    ChatGPT & 67.98 & 84.21 & 44.27 & 58.03 \\
    Llama-2-13b & 58.50 & 65.93 & 35.18 & 45.88\\
    \bottomrule
    \end{tabular}
    }
    \caption{FAMuS Source Validation (SV) results.\vspace{-5mm}}
    \label{tab:source_val_results}
\end{table}

\begin{table*}
    \centering
   \adjustbox{max width=\linewidth}{
    \begin{tabular}{llcccccccccccc}
    \toprule
    & & \multicolumn{6}{c}{\textbf{Report}} &
        \multicolumn{6}{c}{\textbf{Source}} \\
    & & 
        \multicolumn{3}{c}{$\text{CEAF-RME}_{\phi_3}$}  & \multicolumn{3}{c}{$\text{CEAF-RME}_{a}$} & \multicolumn{3}{c}{$\text{CEAF-RME}_{\phi_3}$} & \multicolumn{3}{c}{$\text{CEAF-RME}_{a}$}\\
        \midrule
     & \textbf{Model} & \textbf{P} & \textbf{R} & $\textbf{F}_1$ & \textbf{P} & \textbf{R} & $\textbf{F}_1$ & \textbf{P} & \textbf{R} & $\textbf{F}_1$ & \textbf{P} & \textbf{R} & $\textbf{F}_1$ \\

    \midrule
    \multirow{6}{*}{$-\text{rb}$} & IterX\textsubscript{gold} & 73.11 & 72.00 & 72.55 & 73.56 & 72.44 & 73.00 & 70.46 & 69.16 & 69.80 & 70.58 & 69.28 & 69.92 \\
    & IterX\textsubscript{gold+pred} & 40.57 & 29.38 & 34.08 & 42.24 & 30.59 & 35.48 & 25.07 & 10.82 & 15.11 & 29.85 & 12.88 & 18.00 \\
    & IterX\textsubscript{pred} & 37.63 & 24.14 & 29.41 & 42.16 & 27.04 & 32.94 & 20.83 & \phantom{0}8.63 & 12.21 & 27.63 & 11.45 & 16.19 \\
    & Longformer-QA & \textbf{43.56} & \textbf{40.14} & \textbf{41.78} & \textbf{56.01} & \textbf{51.61} & \textbf{53.72} & \textbf{25.53} & \textbf{22.21} & \textbf{23.75} & \textbf{38.85} & \textbf{33.80} & \textbf{36.15} \\
    & ChatGPT & 33.67 & 32.00 & 32.81 & 51.28 & 48.73 & 49.97 & 14.00 & 12.77 & 13.36 & 33.31 & 30.39 & 31.78 \\
    & Llama-2-13b-chat & 12.97 & 22.76 & 16.52 & 23.65 & 41.49 & 30.13 & 11.14 & \phantom{0}8.52 & \phantom{0}9.65 & 20.11 & 15.36 & 17.42\\

    \midrule
    \multirow{7}{*}{$+\text{rb}$} & Report Baseline (rb) & - & - & -  & - & - & - & 23.59 & 19.68 & 21.46 & \textbf{47.80} & 39.88 & \textbf{43.48} \\
    & IterX\textsubscript{gold} & - & - & -  & - & - & - & 60.38 & 75.95 & 67.28 & 64.12 & 80.65 & 71.45 \\
    & IterX\textsubscript{gold+pred} & - & - & -  & - & - & -  & 24.43 & 19.56 & 21.73 & 38.47 & 30.82 & 34.22 \\
    & IterX\textsubscript{pred} & - & - & -  & - & - & - & 22.24 & 17.38 & 19.51 & 37.42 & 29.24 & 32.83 \\
    & Longformer-QA & - & - & -  & - & - & -  & \textbf{24.12} & \textbf{25.89} & \textbf{24.97} &38.41 & \textbf{41.24} & 39.77 \\
    & ChatGPT & - & - & -  & - & - & -  & 15.93 & 17.95 & 16.88 & 34.99 & 39.42 & 37.07 \\
    & Llama-2-13b-chat & - & - & -  & - & - & -  & 11.11 & \phantom{0}8.52 & \phantom{0}9.64 & 20.24 & 15.51 & 17.56\\
            
    \bottomrule
    \end{tabular}
    }
    \caption{CEAF-RME scores for CDAE on FAMuS test set. The Report Baseline (rb) predicts the gold \emph{report} arguments as the arguments for the source. IterX and Longformer-QA are fine-tuned on FAMuS. ChatGPT and Llama results are evaluated in the few-shot setting. ``$+$/$-$rb'' indicates whether the model is ensembled with the report baseline (see \S\ref{sec:experiments}). \textbf{Bolded} results are best across models within the same $+$/$-$rb setting that do not have access to gold spans for the target document.}
    \vspace{-1em}
    \label{tab:cross_doc_arg_results}
\end{table*}

\subsection{Source Validation}
Table \ref{tab:source_val_results} presents the performance of the SV models described above, along with that of a majority baseline.\footnote{Recall that our SV dataset is class-balanced.} We find that the lemma baseline exhibits the highest overall precision and accuracy. The high precision suggests that the presence of the trigger's lemma in the source document is a very strong indicator that a source is valid. The high accuracy further shows this to be an effective heuristic for distinguishing valid from invalid sources in the dataset as a whole.

By contrast, the Longformer model achieves the highest recall (87.75\%) and $\text{F}_1$ score (75.77\%). The large (almost 30-point) gap in recall between this model and the lemma baseline implies that many valid source documents give paraphrased descriptions of the report event that the lemma baseline cannot identify. A preference for the higher recall provided by the Longformer model (over the higher precision provided by the lemma baseline) may be desirable insofar as sources incorrectly identified as valid can later be invalidated during CDAE---e.g.\ by detecting large mismatches between arguments found in the source and report.

Lastly, ChatGPT trails both previous models on accuracy and $\text{F}_1$, and even trails the majority baseline on the latter, driven by low recall. Its much higher precision (84.21\%) may suggest an over-reliance on simple lexical (e.g.\ lemma-based) cues. Llama 2 shows a similar pattern, though with lower numbers than ChatGPT across the board.

\subsection{Cross-Document Argument Extraction}
Full CDAE results on the FAMuS test set are shown in \autoref{tab:cross_doc_arg_results}.\footnote{As noted in \S\ref{sec:experiments}, results in \autoref{tab:cross_doc_arg_results} use only the human-annotated argument mentions in the reference. Results with full reference argument coreference clusters (generated by F-COREF) are in \autoref{tab:cross_doc_arg_with_silver_coref_results}.} A key (if unsurprising) theme that emerges is the value of high-quality candidate spans. The IterX\textsubscript{gold} results ablate span extraction and reflect argument labeling performance on gold spans for the target document. Unsurprisingly, these are the best absolute numbers, with CEAF-RME $\text{F}_1$ scores in the high 60s and low 70s. Setting aside the report baseline (rb) ensembles, Longformer-QA shows the best performance among models that do not have access to gold arguments, but even these results consistently trail $\text{F}_1$ scores of IterX\textsubscript{gold} by huge margins.

A second, related theme is the difficulty of CDAE on source documents relative to report documents. All models without access to gold spans (both few-shot and fine-tuned) see a significant drop in performance when moving from report extraction to source extraction: even the smallest such drop ($\text{CEAF-RME}_{\phi_3}$ for Llama) is still almost 7 $\text{F}_1$. This is likely a result of models having to consider a much larger set of candidate arguments in the source to identify a set of \emph{correct} ones that is generally comparable in size to the set of gold report arguments (see \autoref{tab:summary-stats}).

We also note that few-shot results with ChatGPT are surprisingly competitive with those of the fine-tuned models, besting IterX\textsubscript{pred} and IterX\textsubscript{gold+pred} on $\text{CEAF-RME}_{a}$ for both report and source extraction (with and without ensembling), and trailing Longformer-QA on the same metric by only a few points in both settings.\footnote{Margins are larger when looking at $\text{CEAF-RME}_{\phi_3}$, as ChatGPT often struggles to recover \emph{exactly} the same mention as is annotated in the reference.}

While the report baseline, which predicts the set of gold arguments from the \emph{report}, is not technically comparable with models in the $-$rb setting, it posts better results than any models in that setting, IterX\textsubscript{gold} excepted. Ensembling other models with the report baseline ($+$rb) tends to substantially improve recall for those models (and sometimes even precision as well), though only the ensembled Longformer-QA actually improves upon the report baseline itself on $\text{CEAF-RME}_{\phi_3}$---and even then, still trailing it on $\text{CEAF-RME}_a$. These results suggest that it is difficult for these models to accurately provide information about the situation that monotonically increases the information that could be obtained from the report alone. Further, the report baseline itself is not particularly high, suggesting that there is a substantial amount of information to be gained about the event by looking at the source; these models are just not able to access it. 

Finally, we note that the generally large absolute differences between $\text{CEAF-RME}_{\phi_3}$ and $\text{CEAF-RME}_{a}$ results for the same model and settings suggest that may predicted arguments are at least partially correct, but do not receive credit under exact match. These results point to the additional information about model performance that incorporating partial span matching into existing metrics can provide for argument extraction. Caution is warranted here though: weakening the requirement for exact matches increases the possibility that models get credit for mentions of incorrect referents---e.g. getting credit for responding \emph{New York} when the correct mention is \emph{New York Times}. Future work on incorporating partial matching into these metrics might investigate using coreference information to penalize models in these cases.

\subsection{Model Performance \& Document Length}
\begin{figure}[t]
    \centering
    \tiny
    \includegraphics[width=\columnwidth, scale=1]{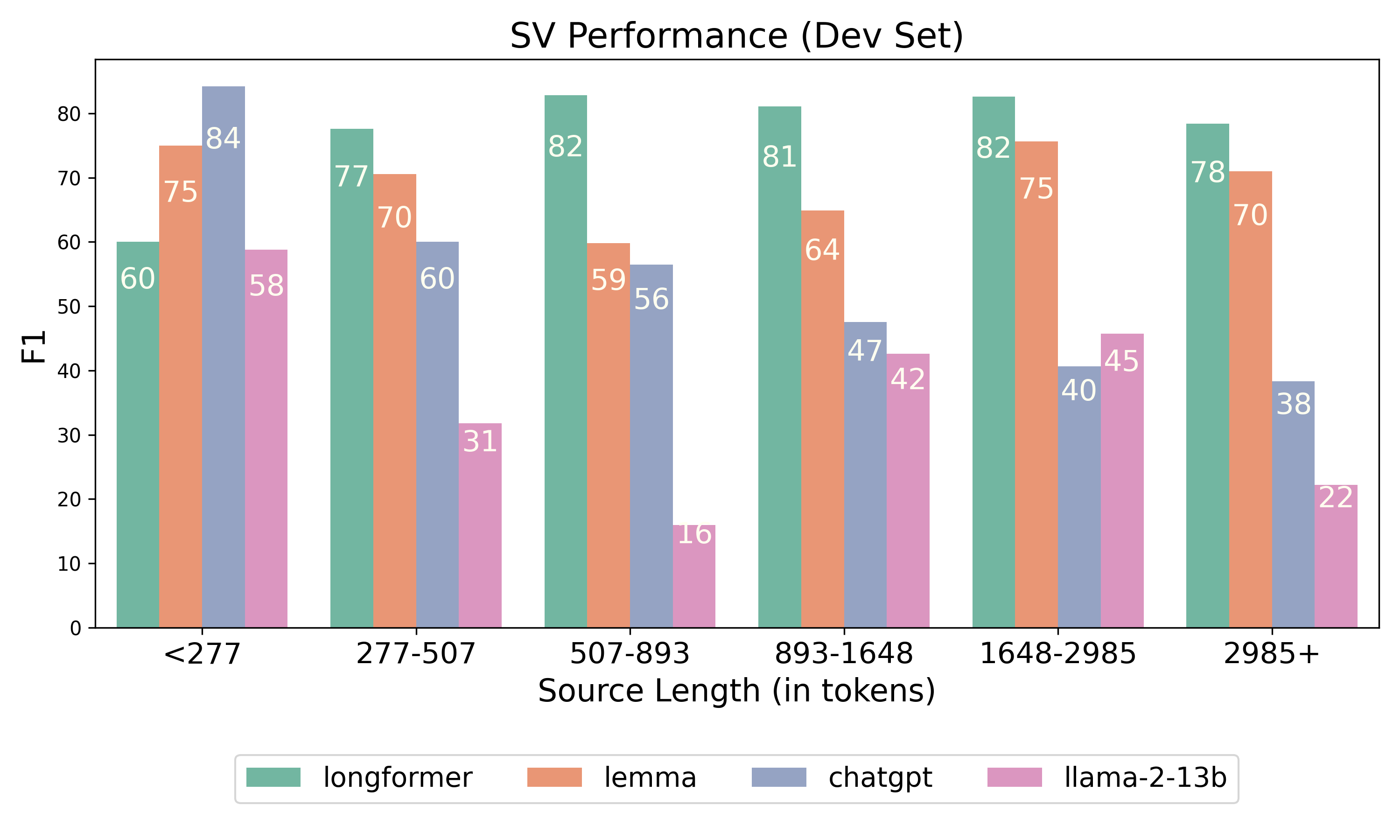}
    \caption{Source Validation $\text{F}_1$ on the FAMuS dev set, broken down by source document length percentile (0-10\%, 10-25\%, 25-50\%, 50-75\%, 75-90\%, 90-100\%).}
    \label{fig:source_val_by_source_length}
\end{figure}
Next, we consider how model performance changes on both tasks as a function of the length of the source document. \autoref{fig:source_val_by_source_length} shows dev set source validation performance of models reported in \autoref{tab:source_val_results}, broken down by source length percentile. Several observations stand out. For one, ChatGPT performs exceptionally well on the shortest documents, achieving 84 $\text{F}_1$ and actually outperforming both the lemma baseline (75 $\text{F}_1$) and Longformer (60 $\text{F}_1$) by wide margins. Across the remaining bins, however, ChatGPT's performance decreases monotonically, faring worse than either of these models, suggesting its strong few-shot capabilities on this task (see above) may be limited to shorter texts. By contrast, Longformer exhibits remarkable consistency across source documents of different lengths: while its performance trails ChatGPT and the lemma baseline on the shortest documents, it outperforms them on all bins of greater length, sustaining $\text{F}_1$ scores between 77 and 82. Llama 2 exhibits the most \emph{in}consistent performance, showing wide variation across bins. 

\begin{figure}[t]
    \centering
    \tiny
    \includegraphics[width=\columnwidth, scale=1]{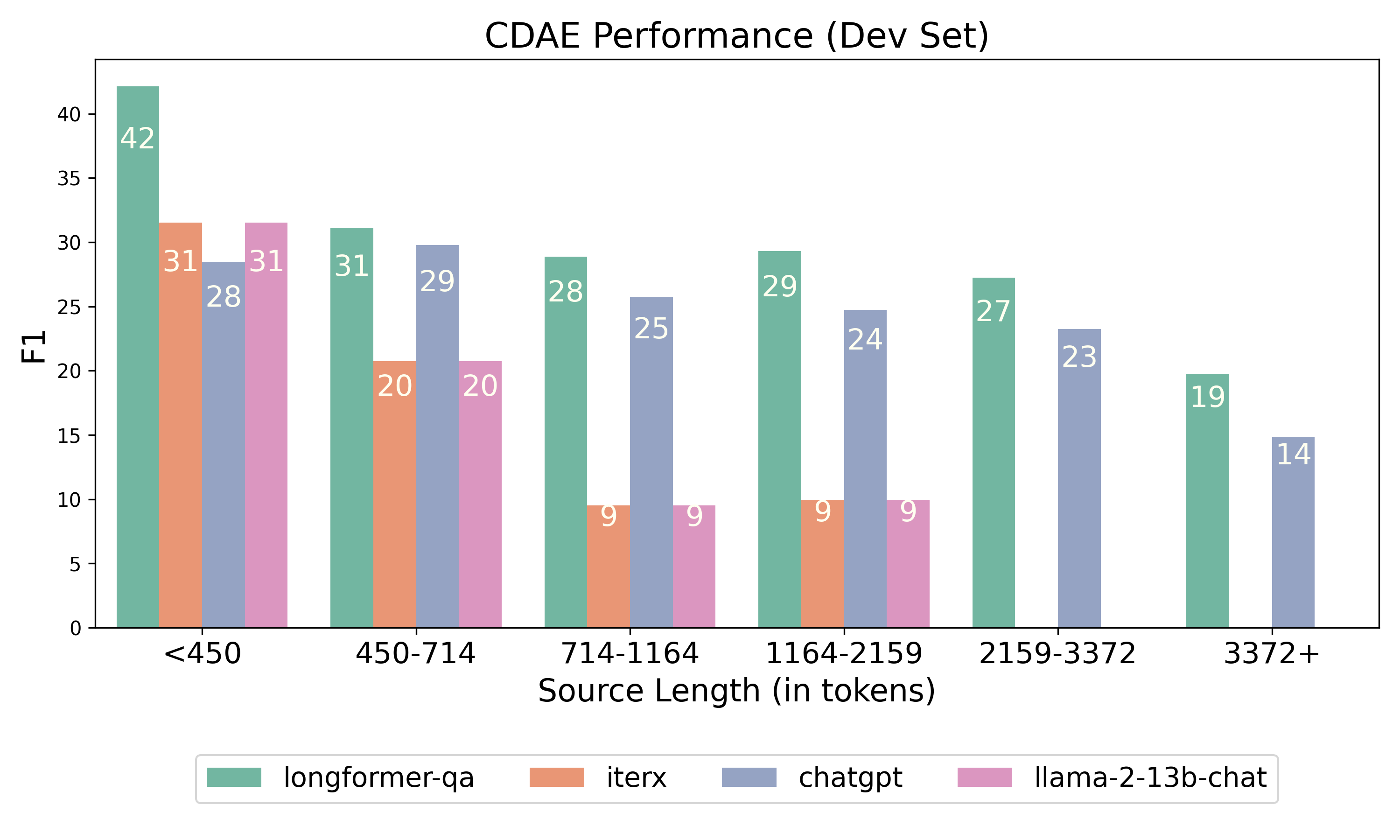}
    \caption{$\text{CEAF-RME}_a$ $\text{F}_1$ on Cross-Document Argument Extraction on source documents, broken down by document length. Percentile bins are the same as in \autoref{fig:source_val_by_source_length}. IterX=IterX\textsubscript{pred} (see \autoref{tab:cross_doc_arg_results}).}
    \vspace{-3em}
    \label{fig:cdae_by_source_length}
\end{figure}

Results for CDAE are shown in \autoref{fig:cdae_by_source_length}. Here, in contrast to the SV results, we observe monotonic or near-monotonic non-increasing performance for all models in moving from shorter to longer documents. This trend is particularly striking for IterX and Llama 2, for which $\text{CEAF-RME}_a$ $\text{F}_1$ scores fall below 10 on source documents in the 25\textsuperscript{th} percentile or greater, and actually fall to 0 for documents in the 75\textsuperscript{th} percentile and above. While ChatGPT and Longformer-QA fare somewhat better, $\text{F}_1$ scores are still under 20 for documents in the 90\textsuperscript{th} percentile and above. This highlights the significant need for argument extraction models that are more robust on long texts.


\section{Conclusion}
\label{sec:conclusion}
We have presented FAMuS, a new dataset comprising \emph{reports} (Wikipedia passages) that describe an event, along with \emph{source} documents for those events---featuring high-quality, full-document FrameNet frame and role annotations on both. We have also introduced two event understanding tasks enabled by FAMuS: \emph{source validation}---determining whether a candidate document is a valid source for a given report event---and \emph{cross-document argument extraction}---extracting the arguments of an identified report event in both the report and its source. We have provided baselines for both tasks, along with detailed analysis of their performance, and release both these models and our data to facilitate future research.

\section*{Limitations}
\label{sec:limitations}
One limitation of FAMuS is that its annotations are \emph{non-exhaustive}: only the arguments of the (single) target event are annotated in the report and source. This makes it unsuited to training models for full (document-level) event extraction, in which systems typically may have to extract multiple events. Remedying this shortcoming is one of our primary goals for follow-up work.

Additionally, while FAMuS provides annotations for argument coreference, these are model-predicted, and thus will contain some noise. (Granted, this is irrelevant for evaluation against only the gold annotated spans, as in \autoref{tab:cross_doc_arg_results}.)

Finally, because the valid source documents in FAMuS are cited by their corresponding reports, this may result in artificially high agreement between the arguments in the report and those in the source. Different internet sources routinely give somewhat differing, and even conflicting, accounts of the same event, and insofar as Wikipedia articles overwhelmingly cite documents \emph{in support} of the claims they make, FAMuS likely overestimates the level of inter-document consensus present on the internet more broadly.


\section*{Ethics Statement}
\label{sec:ethics}
We do not believe this work raises significant ethical issues.

\section*{Acknowledgements}
\label{sec:acknowledgments}
This work was supported by DARPA AIDA, IARPA BETTER, and NSF-BCS (2040831). The views and conclusions contained in this work are those of the authors and should not be interpreted as necessarily representing the official policies, either expressed or implied, or endorsements of DARPA, IARPA, or the U.S.\ Government. The U.S.\ Government is authorized to reproduce and distribute reprints for governmental purposes notwithstanding any copyright annotation therein.

\bibliography{anthology,custom}

\clearpage
\appendix

\section{IAA and Annotation Correction}
\label{app:agreement}
This appendix offers additional details on annotator agreement and annotation correction for CDAE.
\subsection{Agreement}
Here, we describe the agreement metric used for computing inter-annotator agreement (IAA) for CDAE annotation. We compute a $\text{F}_1$ score based on the maximum normalized edit distance ($a$) between annotated and reference argument \emph{mentions} given in Eq.\ (\ref{eq:a}). If $r$ is a role in the role set $R_f$ for a frame $f$; $p_r$ is a predicted mention; $g_r$ is a reference mention; $C_{g_r}$ is the reference entity containing mention $g_r$; and $\epsilon$ is the ``null'' span (indicating the absence of an argument), we compute this $\text{F}_1$ score based on the following counts of true positive (TP), false positive (FP), and false negative (FN) arguments:
\begin{flalign}
\begin{aligned}
  \operatorname{TP} =  
  \sum_{\substack{C_{g_r} \ne \phi \; \cap \; p_r \ne \epsilon \\ r \in R_f}} a 
\end{aligned} && \notag
\end{flalign}
\begin{flalign}
\begin{aligned}
\operatorname{FP} =
\sum_{\substack{C_{g_r} \ne \phi \; \cap \; p_r \ne \epsilon \\  r \in R_f}}
\frac{1-a}{2}
\end{aligned} && \notag + && \begin{aligned}
\sum_{\substack{C_{g_r} = \phi \; \cap \; p_r \ne \epsilon \\ r \in R_f}}
\vphantom{\frac{1-a}{2}} 1
\end{aligned} && \notag
\end{flalign}
\vspace{-1.2em}
\begin{flalign}
\begin{aligned}
\operatorname{FN} =
\sum_{\substack{C_{g_r} \ne \phi \; \cap \; p_r \ne \epsilon \\  r \in R_f}}
\frac{1-a}{2}
\end{aligned} && \notag + && \begin{aligned}
\sum_{\substack{C_{g_r} \ne \phi \; \cap \; p_r = \epsilon \\ r \in R_f}}
\vphantom{\frac{1-a}{2}} 1
\end{aligned} && \notag
\end{flalign}

\subsection{Annotation Correction}
\begin{figure}
    \centering
    \includegraphics[width=\columnwidth, scale=1]{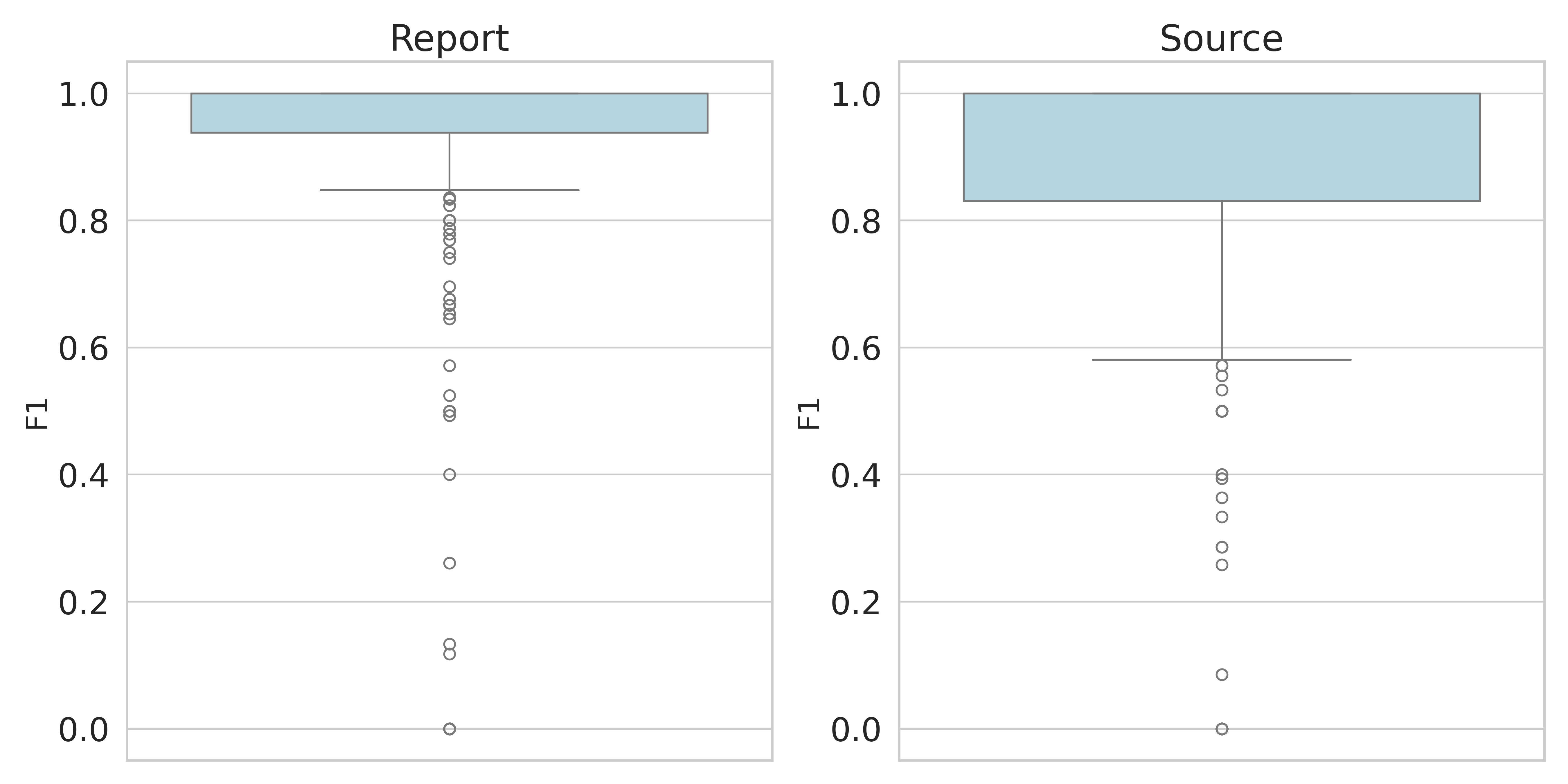}
    \caption{Boxplots for agreement $\text{F}_1$ between bottom quartile report and source CDAE annotations before and after correction by the authors.}
    \label{fig:boxplot_gold_platinum}
\end{figure}

\begin{figure*}
    \centering
    \includegraphics[width=\textwidth]{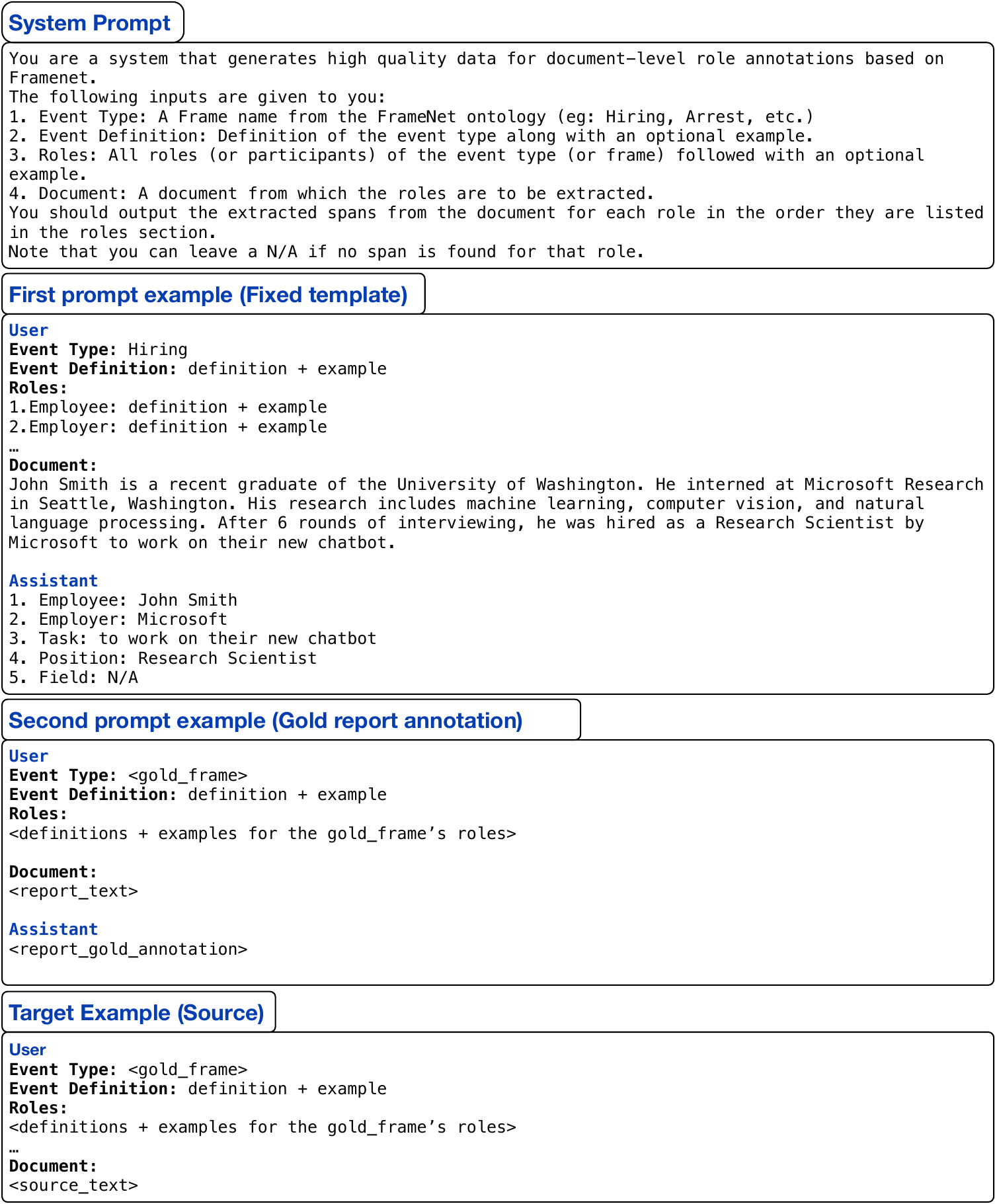}
    \caption{Prompt template used to generate CDAE annotations on the source for annotation correction. Note the use of gold report annotation as the second prompt example.}
    \label{fig:chatgpt_annotation_correction}
\end{figure*}

As discussed in \S\ref{sec:data_collection}, we use the agreement metric above to evaluate the similarity between annotators' CDAE annotations on the source text and those produced by ChatGPT (\texttt{gpt-3.5-turbo-0301}) in order to identify potentially lower quality human annotations. At the end of each round of CDAE annotation, (report, source) pairs for which the source agreement score with ChatGPT falls in the bottom quartile are manually verified and corrected by the authors. The prompt template we use to obtain source document CDAE annotations with ChatGPT is shown in \autoref{fig:chatgpt_annotation_correction}. The prompt includes two examples in the chat history, where the first is the same across report documents, while the second uses the gold annotation from the report associated with the target source document. We set \texttt{max\_tokens} to 128, \texttt{top\_p} to 1.0, and \texttt{temperature} to 0, with no presence or frequency penalties. Figure \ref{fig:boxplot_gold_platinum} shows boxplots of the agreement $\text{F}_1$ scores between the report and source annotations before and after manual correction by the authors, aggregated over all rounds of annotation. Note that the majority of corrected annotations actually exhibit perfect agreement with their uncorrected counterparts, resulting in high mean scores of 0.90 and 0.85 for reports and sources, respectively, and offering compelling evidence for the quality of the annotations overall.

\begin{figure*}
    \centering
    \includegraphics[width=\textwidth]{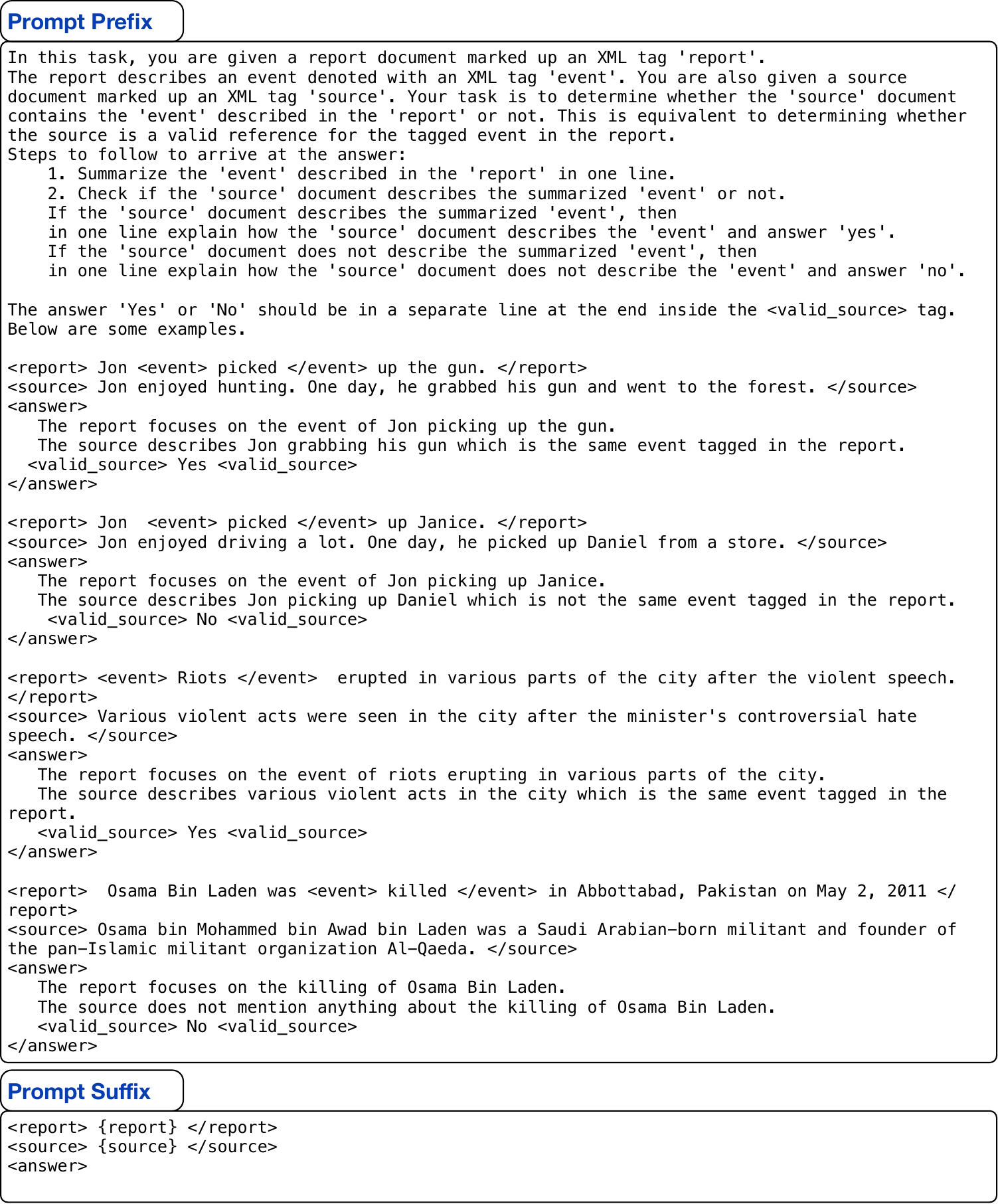}
    \caption{Prompt template used for ChatGPT and Llama 2 on SV.}
    \label{fig:sv_prompt_prefix}
\end{figure*}

\begin{figure*}
    \centering
    \includegraphics[width=\textwidth]{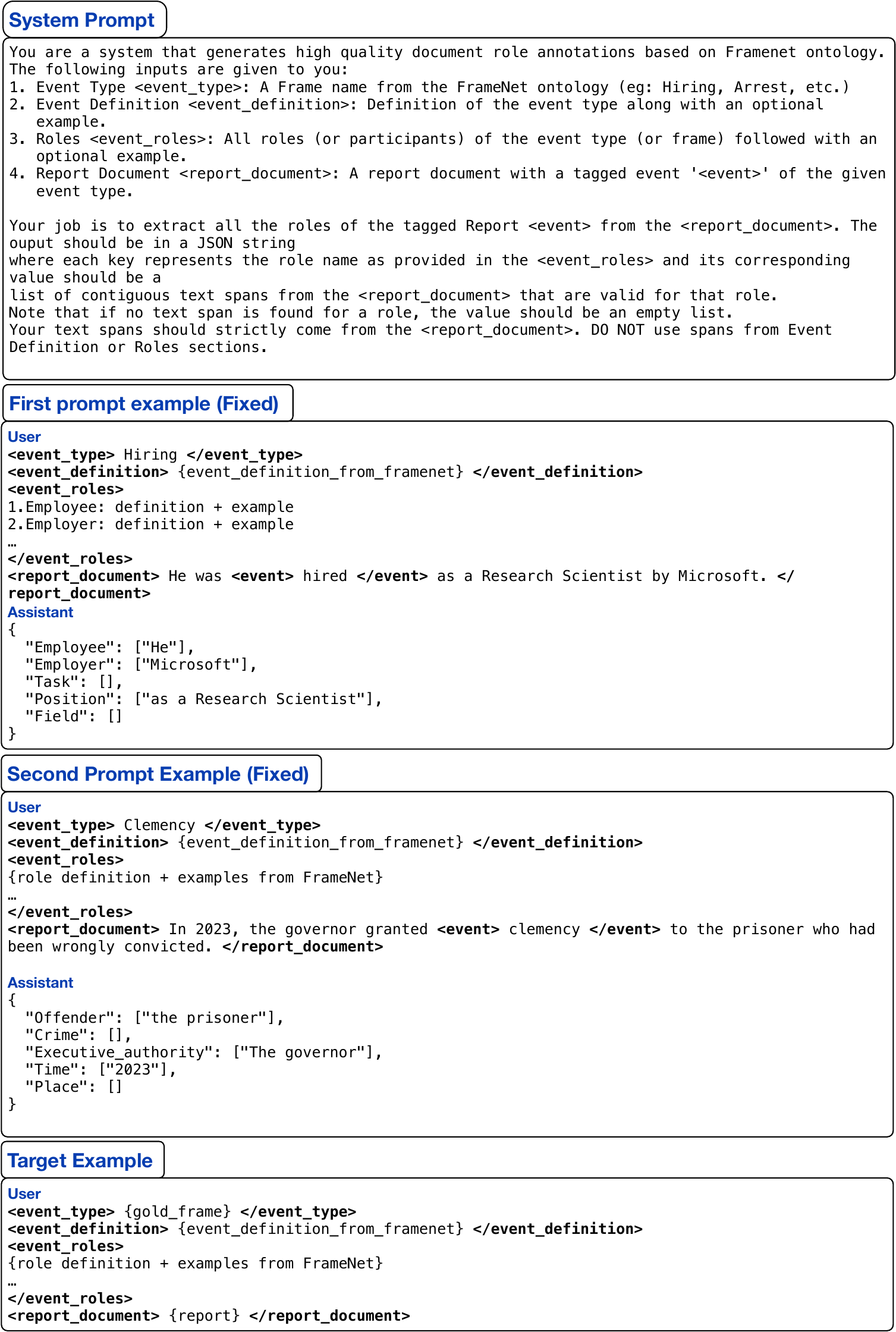}
    \caption{Prompt template used for ChatGPT and Llama 2 on CDAE for report documents.}
    \label{fig:cdae_prompt_report}
\end{figure*}

\begin{figure*}
    \centering
    \includegraphics[width=\textwidth]{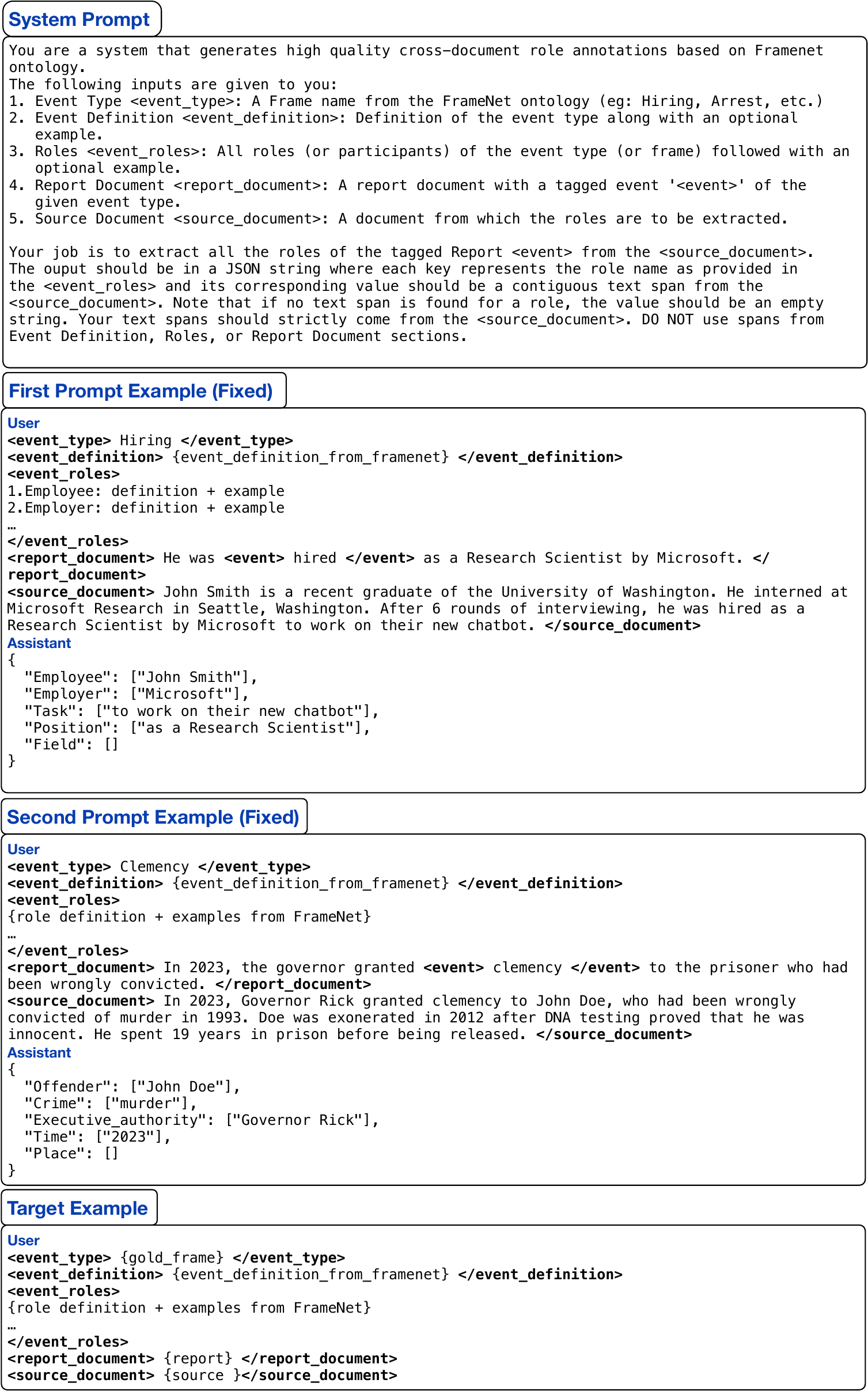}
    \caption{Prompt template used for ChatGPT and Llama 2 on CDAE for source documents.}
    \label{fig:cdae_prompt_source}
\end{figure*}

\section{Model Details}
\label{app:model_details}
This appendix presents model implementation details, hyperparameters, and prompts. Further details are available in the GitHub repo (\url{https://github.com/FACTSlab/FAMuS}).

\subsection{Source Validation}

\paragraph{Longformer} We use the LongformerForSequenceClassification class from the HuggingFace Transformers library \citep{wolf-etal-2020-transformers} to fine-tune Longformer for source validation. We fine-tune for 15 epochs with a batch size of 2, and an initial warmup phase of 400 steps. We conduct limited hyperparameter search using Optuna \citep{akiba2019optuna}, targeting the learning rate and weight decay, and varying them logarithmically from 1e-6 to 1e-4 and from 1e-6 to 1e-1 respectively. This process is conducted over 5 trials, with the optimal setting selected based on the highest validation accuracy. We then fine-tune a final Longformer model for 30 epochs using the best hyperparameter configuration, using the checkpoint with highest dev accuracy across all 30 epochs for the final evaluation.

\paragraph{ChatGPT}
We use \texttt{gpt-3.5-turbo-0301} and do not perform any fine-tuning or hyperparameter search, evaluating only in the few-shot setting. \autoref{fig:sv_prompt_prefix} presents a sketch of the prompt we use. We set \texttt{max\_tokens} to 128, \texttt{top\_p} to 1.0, and \texttt{temperature} to 0, with no presence or frequency penalties.

\paragraph{Llama 2}
We use \texttt{llama-2-13b} for source validation and do not perform any fine-tuning or hyperparameter search (just as with ChatGPT). We use the default hyperparameters, except for \texttt{max\_seq\_len} (5,000), \texttt{max\_gen\_len} (128), \texttt{top\_p} (0.9), and \texttt{temperature} (0.0). The prompt is the same as the one used for ChatGPT on SV (\autoref{fig:sv_prompt_prefix}).

\subsection{Cross-Document Argument Extraction}

\paragraph{Longformer}
We use the LongformerForQuestionAnswering class from the HuggingFace transformers library to fine-tune the Longformer-QA model on the recasted CDAE datasets for both report and source. We fine-tune for a maximum of 10 epochs with a batch size of 1. As with Longformer for SV, we use the Optuna library for hyperparameter tuning to optimize the learning rate and weight decay, varying them logarithmically from 1e-6 to 1e-4 and 1e-6 to 1e-1 respectively. This process is conducted over 5 trials, with the optimal setting selected based on the lowest validation loss.

\paragraph{IterX}
We base our IterX hyperparameters on the best ones reported for the MUC-4 task in Table 6 of \citet{chen-etal-2023-iterative}, though with two important differences. First, as noted in \S\ref{sec:experiments}, the CDAE task requires extraction of only a single template per document. As such, we set the maximum number of templates to decode (``\#Max Iterations'') to 1. Second, \citeauthor{chen-etal-2023-iterative} train their model for MUC-4 on \emph{predicted} spans only, whereas we use different sets of spans for training depending on the setting (\textbf{gold}, \textbf{predicted}, or \textbf{gold and predicted}). All models are trained for a maximum of 150 epochs with a patience of 30, using $\text{CEAF-RME}_{\phi_3}$ on the dev set as the validation metric. Our implementation is available at the following, lightly adapted fork of the public IterX repo: \url{https://github.com/sidsvash26/iterx/tree/sv/famus}.

\paragraph{ChatGPT}
As with SV, we do not perform any fine-tuning or hyperparameter search on ChatGPT for CDAE. The model version (\texttt{gpt-3.5-turbo-0301}) and hyperparameters used here are also identical to those used for ChatGPT on SV. We use separate prompts for extraction on report and source documents. The prompt used for source documents is sketched in \autoref{fig:cdae_prompt_source}. It consists of a system prompt, two example extractions (included in the chat history), followed by the target example on which extraction is to be performed. Note that this is different from the prompt used to generate CDAE annotations for purposes of annotation correction (\autoref{fig:chatgpt_annotation_correction}).

\paragraph{Llama 2}
We use \texttt{llama-2-13b-chat} in lieu of \texttt{llama-2-13b} (used in SV), though all hyperparameters are the same as those used for Llama on SV. The prompt is the same as that used for ChatGPT for CDAE.

\section{Frame Selection}
\label{app:frame_selection}
We follow the frame selection methodology of \citet{barham2023megawika} for selecting situation-denoting frames. Drawing inspiration from \citet{moens-steedman-1988-temporal}, we focus on the top-level \textsc{Event}, \textsc{State}, and \textsc{Process} FrameNet frames. We initially take these three frames and all those related to them via the \textsc{inheritance}, \textsc{subframe}, or \textsc{precedes} relations, on the assumption that the set of situation-denoting frames is closed under these relations, yielding 387 frames.\footnote{See the FrameNet Lattice List: \url{https://framenet.icsi.berkeley.edu/FrameLatticeList}}

However, some of these frames also inherit from other top-level frames that are \emph{not} situation-denoting (i.e.\ \textsc{Relation}, \textsc{Entity}, and \textsc{Locale}). We remove all such frames from the set above, which leaves 369 frames remaining.

Finally, because we are reliant on an existing FrameNet parser for data collection \citep{xia-etal-2021-lome}, we must further subset to those frames for which there is FrameNet training data and that therefore exist in the model's vocabulary. This yields the final set of 328 frames reported in \S\ref{sec:data_collection}.

\section{Additional Statistics}
\label{app:dataset_statistics}

\subsection{Similarity Between Report and Source}
\begin{figure}
    \centering
    \includegraphics[width=\columnwidth, scale=1]{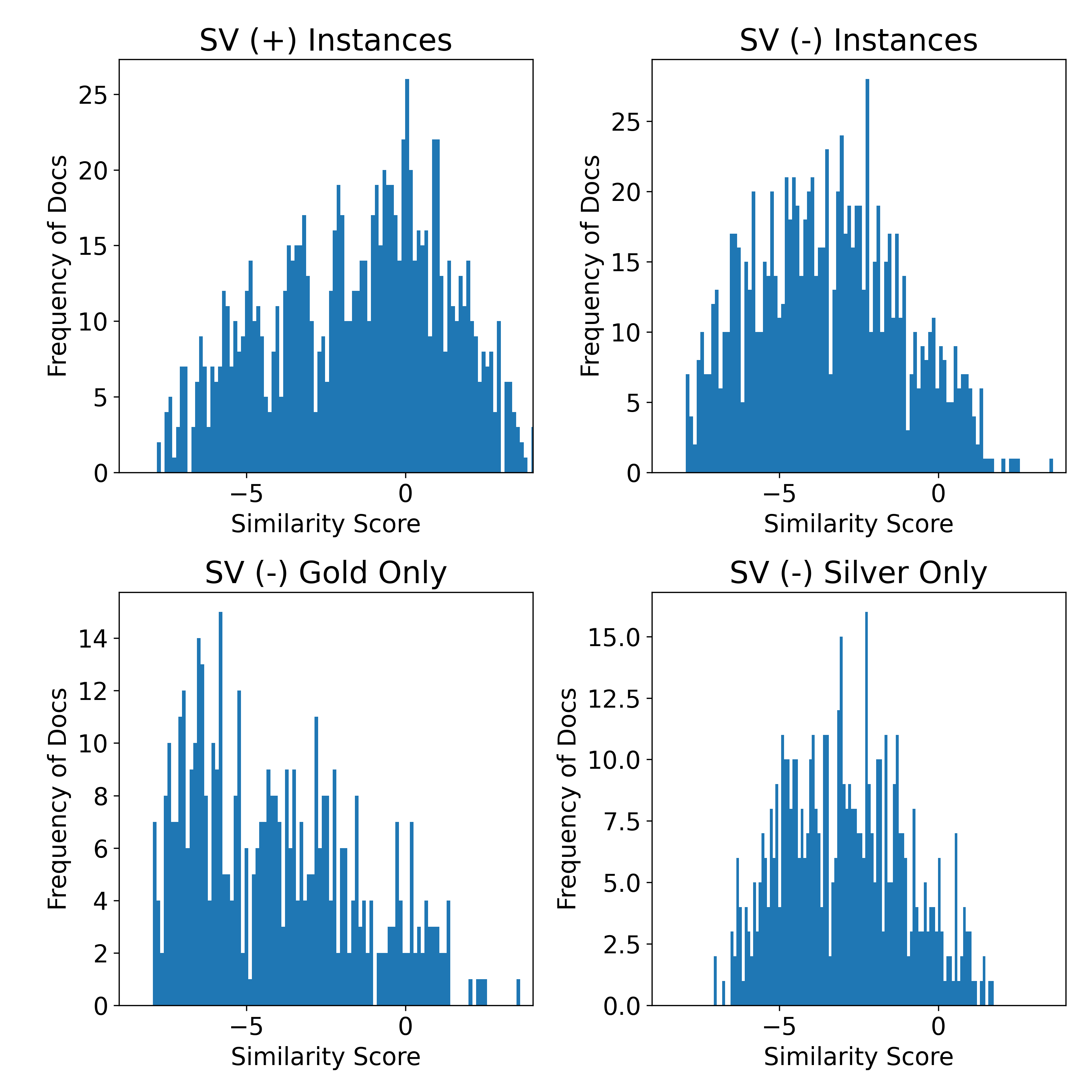}
    \caption{Histogram of SimLM similarity scores between the Report and the Source Text across SV train and dev examples.}
    \label{fig:histogram_simLM}
\end{figure}

As discussed in \S\ref{sec:data_collection}, we use SimLM to compute the similarity between (report, source) pairs when automatically generating negative examples for the SV task. \autoref{fig:histogram_simLM} presents the distributions of the SimLM scores for all positive SV examples (top left), all negative SV examples (top right), human-annotated (gold) negative examples (bottom left), and automatically curated (silver) negative examples (bottom right). As may be expected, the modal similarity in the negative example plots is less than that of the positive examples. This is true even for the silver negative examples, for which we deliberately selected sources based on similarity to a target report, which offers further evidence that we are unlikely to be accidentally including positive source documents in the automatically generated negative examples. 

\subsection{Argument Distances}
\begin{figure}
    \centering
    \includegraphics[width=\columnwidth, scale=1]{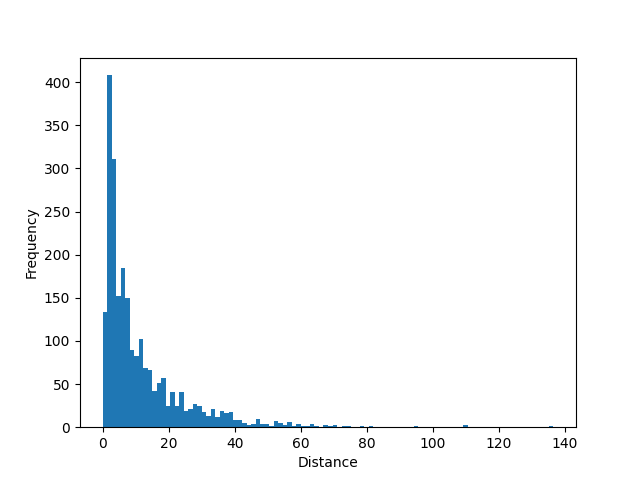}
    \caption{Histogram of word distances between triggers and their arguments in report documents from train.}
    \label{fig:trigger-distance}
\end{figure}

\begin{figure}
    \centering
    \includegraphics[width=\columnwidth]{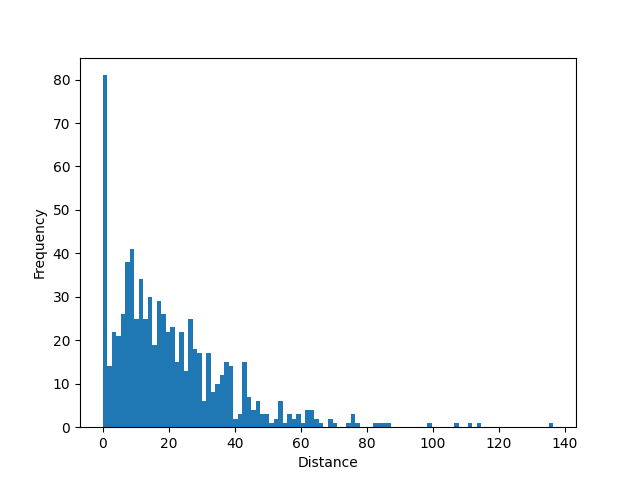}
    \includegraphics[width=\columnwidth]{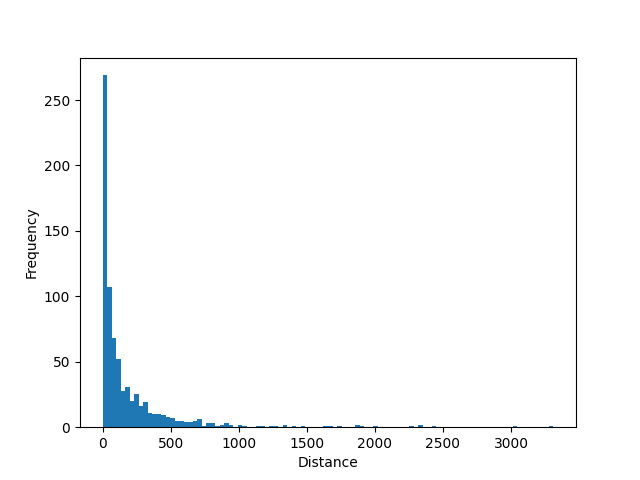} 
    \caption{Histogram of word distances between the first and last arguments in \emph{report} documents (top) and source documents (bottom) from the train split.}
    \label{fig:arg-distances}
\end{figure}

Here, we report distributions and statistics for word distances between (1) event triggers and their arguments in training split report documents (\autoref{fig:trigger-distance}); and (2) the first and last arguments annotated in each report and source document (\autoref{tab:role-distance}, \autoref{fig:arg-distances}) in the training split. Recall that in contrast to a number of resources for event argument extraction \citep[EAE;][]{ebner-etal-2020-multi, li-etal-2021-document}, FAMuS permits arguments to be annotated \emph{anywhere} in the report and source documents. 

\begin{table}[]
    \centering
    \adjustbox{max width=\columnwidth}{
    \begin{tabular}{ll|rr}
    \toprule
         & & Train & Dev \\
    \midrule
        & Mean & 20.5 & 25.6 \\
        Report & Median & 16.5 & 22.0\\
        & Std Dev & 18.2 & 18.9 \\
        \midrule
        & Mean  & 193.7 & 310.4 \\
        Source & Median & 67.3 & 122.0\\
        & Std Dev & 353.6 & 529.0 \\
    \bottomrule
    \end{tabular}
    }
    \caption{Statistics for word distances between the first and last arguments in report and source documents.}
    \label{tab:role-distance}
\end{table}

\section{Additional Results}
\label{app:additional_results}
\autoref{tab:cross_doc_arg_with_silver_coref_results} presents CEAF-RME scores on the same models as in \autoref{tab:cross_doc_arg_results}, but using the full coreference cluster for each gold argument (as predicted by F-COREF) in the metric computation. The results are qualitatively similar (Longformer-QA remains dominant in the $-$rb setting and the Report Baseline still generally outperforms models in the $+$rb setting), though absolute $\text{F}_1$ scores are noticeably reduced. This is driven by lower recall, likely due to a combination of (1) models tending to predict only a single mention per (gold) entity (2) possibly noisy reference entities, owing to their being model-generated.

\begin{table*}
\centering
   \adjustbox{max width=\linewidth}{
    \begin{tabular}{llcccccccccccc}
    \toprule
    & & \multicolumn{6}{c}{\textbf{Report}} &
        \multicolumn{6}{c}{\textbf{Source}} \\
    & & 
        \multicolumn{3}{c}{$\text{CEAF-RME}_{\phi_3}$}  & \multicolumn{3}{c}{$\text{CEAF-RME}_{a}$} & \multicolumn{3}{c}{$\text{CEAF-RME}_{\phi_3}$} & \multicolumn{3}{c}{$\text{CEAF-RME}_{a}$}\\
        \midrule
     & \textbf{Model} & \textbf{P} & \textbf{R} & $\textbf{F}_1$ & \textbf{P} & \textbf{R} & $\textbf{F}_1$ & \textbf{P} & \textbf{R} & $\textbf{F}_1$ & \textbf{P} & \textbf{R} & $\textbf{F}_1$ \\

    \midrule
    \multirow{6}{*}{$-$rb} & IterX\textsubscript{gold} & 73.11 & 57.17 & 64.17 & 73.56 & 57.53 & 64.56 & 70.46 & 30.26 & 42.34 & 70.61 & 30.33 & 42.43  \\
    & IterX\textsubscript{gold+pred} & 40.95 & 23.55 & 29.90 & 42.24 & 24.29 & 30.84 & 27.47 & \phantom{0}5.19 & \phantom{0}8.73 & 31.63 & \phantom{0}5.97 & 10.05 \\
    & IterX\textsubscript{pred} & 38.06 & 19.39 & 25.69 & 42.33 & 21.56 & 28.57 & 23.61 & \phantom{0}4.28 & \phantom{0}7.25 & 29.80 & \phantom{0}5.40 & \phantom{0}9.15 \\
    & Longformer-QA & \textbf{44.31} & \textbf{32.42} & \textbf{37.44} & \textbf{56.92} & \textbf{41.65} & \textbf{48.10} & \textbf{29.10} & \textbf{11.08} & \textbf{16.05} & \textbf{41.86} & \textbf{15.93} & \textbf{23.08} \\
    & ChatGPT & 35.85 & 27.05 & 30.84 & 53.27 & 40.20 & 45.82 & 17.28 & \phantom{0}6.90 & \phantom{0}9.86 & 36.48 & 14.56 & 20.82 \\
    & Llama-2-13b-chat & 13.68 & 19.06 & 15.93 & 24.20 & 33.72 & 28.18 & 12.35 & \phantom{0}4.13 & \phantom{0}6.19 & 21.41 & \phantom{0}7.16 & 10.73  \\

    \midrule
    \multirow{6}{*}{$+$rb} & Report Baseline (rb) & - & - & - & - & - & - & \textbf{28.69} & 10.47 & 15.34 & \textbf{51.08} & 18.65 & \textbf{27.32} \\
    & IterX\textsubscript{gold} & - & - & - & - & - & - & 61.21 & 33.69 & 43.46 & 64.91 & 35.72 & 46.09 \\
    & IterX\textsubscript{gold+pred} & - & - & - & - & - & - & 27.16 & \phantom{0}9.52 & 14.09 & 40.55 & 14.21 & 21.05 \\
    & IterX\textsubscript{pred} & - & - & - & - & - & - & 25.04 & \phantom{0}8.56 & 12.76 & 39.93 & 13.65 & 20.35 \\
    & Longformer-QA & - & - & - & - & - & - & 26.90 & \textbf{12.64} & \textbf{17.20} & 41.30 & \textbf{19.40} & 26.40 \\
    & ChatGPT & - & - & - & - & - & - & 19.10 & \phantom{0}9.42 & 12.61 & 38.24 & 18.85 & 25.26 \\
    & Llama-2-13b-chat  & - & - & - & - & - & - & 12.31 & \phantom{0}4.13 & \phantom{0}6.18 & 21.44 & \phantom{0}7.19 & 10.77 \\
    \bottomrule
    \end{tabular}
    }
    \caption{Results for the same models as reported in \autoref{tab:cross_doc_arg_results}, but using full (F-COREF-predicted) coreference clusters for the reference arguments when computing CEAF-RME scores.}
    \vspace{-1em}
    \label{tab:cross_doc_arg_with_silver_coref_results}
\end{table*}

\end{document}